\documentclass[journal]{IEEEtran}
\usepackage{cite}
\usepackage{url}

\usepackage[usenames, dvipsnames]{color}

\ifCLASSINFOpdf
  \usepackage[pdftex]{graphicx}

\else
\fi
\usepackage{amsfonts}
\usepackage{amsmath}

\usepackage{booktabs}
\usepackage{multirow}
\usepackage{colortbl}
\usepackage{newclude}
\usepackage{authblk}
\usepackage{tablefootnote}
\usepackage{threeparttable}
\usepackage[colorlinks,linkcolor=blue]{hyperref}

\usepackage{stfloats}

\newcommand{\wq}[1]{\textcolor{black}{#1}}
\newcommand{\ym}[1]{\textcolor{black}{#1}}
\newcommand{\bb}[1]{\textcolor{black}{#1}}
\newcommand{\ld}[1]{\textcolor{black}{#1}}

\usepackage[switch]{lineno}
\begin{document}
%
\title{Open World Object Detection: A Survey}
%
%
%
\author{Yiming~Li, Yi~Wang, Wenqian~Wang, Dan~Lin, Bingbing~Li, Kim-Hui~Yap
\thanks{This study is supported under the RIE2020 Industry Alignment Fund – Industry Collaboration Projects (IAF-ICP) Funding Initiative, as well as cash and in-kind contribution from the industry partner(s).}

\thanks{Yiming Li and Kim-Hui Yap are with the School of Electrical and Electronic Engineering, Nanyang Technological University, Singapore. email: yiming008@e.ntu.edu.sg, ekhyap@ntu.edu.sg. \textit{(Corresponding author: Kim-Hui Yap)}}

\thanks{Yi Wang is with the Department of Electrical and Electronic Engineering, the Hong Kong Polytechnic University, Hong Kong SAR. email: yi-eie.wang@polyu.edu.hk.}

\thanks{Wenqian Wang and Dan Lin are with the Continental-NTU Corporate Lab, Nanyang Technological University, Singapore. email: wenqian.wang, dan.lin@ntu.edu.sg.}

\thanks{Bingbing Li is with the Continental Automotive Singapore Pte. Ltd. email: bingbing.li@continental-corporation.com.}
}
\maketitle

\begin{abstract}
Exploring new knowledge is a fundamental human ability that can be mirrored in the development of deep neural networks, especially in the field of object detection. Open world object detection (OWOD) is an emerging area of research that adapts this principle to explore new knowledge. It focuses on recognizing and learning from objects absent from initial training sets, thereby incrementally expanding its knowledge base when new class labels are introduced. This survey paper offers a thorough review of the OWOD domain, covering essential aspects, including problem definitions, benchmark datasets, source codes, evaluation metrics, and a comparative study of existing methods. Additionally, we investigate related areas like open set recognition (OSR) and incremental learning (IL), underlining their relevance to OWOD. Finally, the paper concludes by addressing the limitations and challenges faced by current OWOD algorithms and proposes directions for future research. To our knowledge, this is the first comprehensive survey of the emerging OWOD field with over one hundred references, marking a significant step forward for object detection technology. A comprehensive source code and benchmarks are archived and concluded at \href{https://github.com/ArminLee/OWOD_Review}{https://github.com/ArminLee/OWOD\_Review}.
\end{abstract}

%
\IEEEpeerreviewmaketitle

\section{Introduction}
\IEEEPARstart{O}{bject} detection\wq{, i.e., locating and identifying objects in images, is crucial for real-world applications.
Object detection can be utilized in autonomous driving to identify and respond to obstacles, and in robot vision to navigate and interact with objects. Besides, there are many application scenarios, e.g., video surveillance to monitor activities, medical imaging to detect anomalies, and industrial automation to ensure quality control.
}
However, \wq{traditional object detection works \cite{redmon2016yolo9000, FasterRCNN,yu20201st,zhu2023evidential} assume} that all classes to be detected are present during training, which leads to two issues: 1) An image may contain objects from unknown classes that object detectors should classify; 2) When information about these unknowns becomes available, the models should be able to learn the novel classes incrementally without forgetting the learned classes.

\begin{figure}
\centering
\includegraphics[scale=0.4]{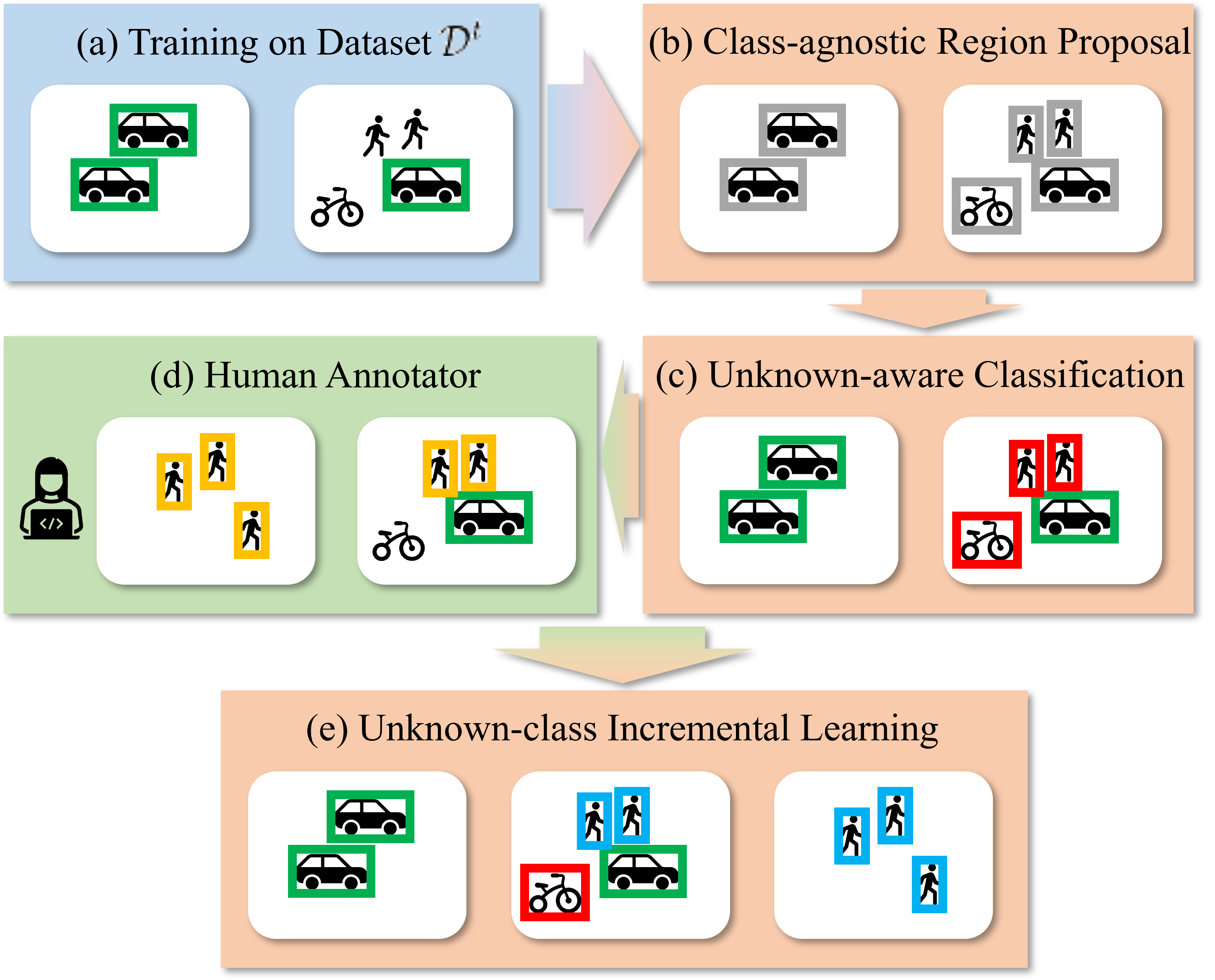}\caption{Demonstration of Open World Object Detection problem. (a) Training on dataset $\mathcal{D}^{t}$. The training on dataset $\mathcal{D}^{t}$ at time $\textit{t}$ adopts the annotation of known categories shown in \textcolor[RGB]{0,176,80}{\textbf{green}} bounding boxes. (b) Class-agnostic Region Proposal. The OWOD model will first generate \textcolor[RGB]{166,166,166}{\textbf{gray}} bounding box proposals for all classes of objects, as shown in the Unknown Proposal process. (c) Unknown-aware Classification. The previously known and unknown categories will be detected separately as \textcolor[RGB]{0,176,80}{\textbf{green}} and \textcolor[RGB]{255,0,0}{\textbf{red}} bounding boxes. \ym{(d) Human Annotator. New annotations for novel objects such as `human' shown as \textcolor[RGB]{255,192,0}{\textbf{yellow}} bounding boxes will be introduced to the model by manual annotation or auto-labeling techniques. The `bicycle' category is not annotated for the illustrated setting.} (e) Unknown-class Incremental Learning. After the incremental learning process of novel categories, the OWOD model is able to detect all the previously known, novel, and unknown object proposals as \textcolor[RGB]{0,176,80}{\textbf{green}}, \textcolor[RGB]{0,176,240}{\textbf{blue}}, and \textcolor[RGB]{255,0,0}{\textbf{red}} bounding boxes. \ym{The newly introduced `human' category will be recognized as `human', but the unlabeled `bicycle' category will still be detected as `unknown' class.} (Best viewed in color)}\label{Demo}\vspace{-0.4cm}
\end{figure}

\wq{Inspired by empirical investigations \cite{wisdom, Curious} in developmental psychology, which revealed that recognizing gaps in one's knowledge is crucial for fostering curiosity and the desire to acquire new knowledge \cite{children, curiousMind}, Joseph \textit{et al.} \cite{ORE} first propose the concept of Open World Object Detection (OWOD). This approach focuses on detecting both known and unknown classes while incrementally learning the identified unknown ones, reflecting the dynamic nature of the real world where knowledge continuously grows.}


OWOD is emerging and receiving increasing attention in recent research works \cite{OW-DETR, PROB,2B-OCD,CAT,OW-RCNN} as it differs fundamentally from traditional object detection. Traditional methods are restricted to a fixed set of predefined classes, limiting their adaptability in dynamic environments. In contrast, OWOD is designed for continuous adaptability, acknowledging the unpredictability of real-world scenarios. While traditional models may misclassify or overlook unfamiliar objects, OWOD can identify and potentially flag such ``unknowns'', emphasizing its adaptability and scalability. Traditional systems, reliant on exhaustive datasets, often require computationally-intensive retraining to accommodate new classes. OWOD, with its incremental learning approach, can seamlessly integrate new data, eliminating the need for such extensive retraining. A notable advantage of OWOD is its mitigation of ``catastrophic forgetting'', a challenge in incremental learning areas where models lose old knowledge when updated with new data. This ensures that OWOD retains previously learned information while adapting to new inputs. In essence, OWOD offers a more adaptive, scalable, and efficient approach to object detection, making it better suited for the dynamic and unpredictable nature of real-world scenarios.

\begin{figure*}
\centering
\includegraphics[scale=0.45]{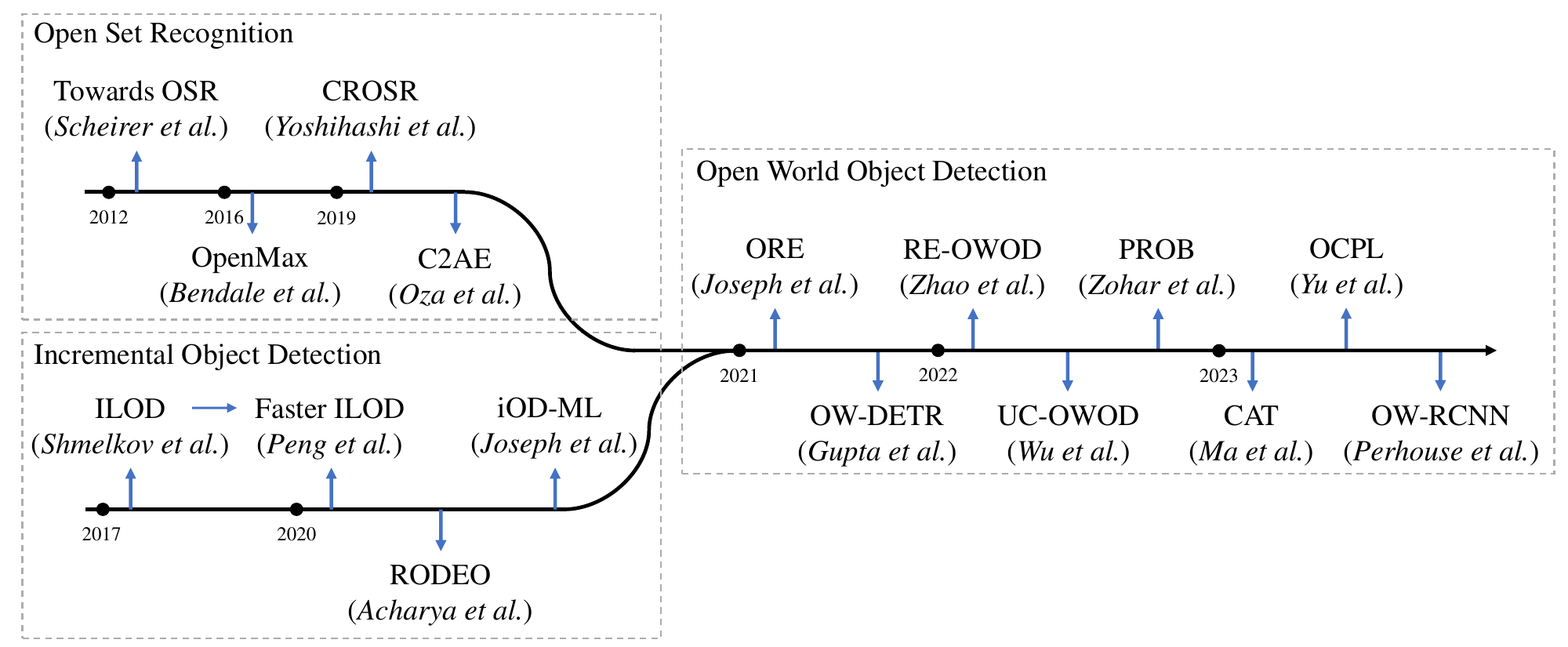}\caption{Milestones and development of open world object detection. \wq{The detailed description of related areas and methods are presented in Section \ref{sec1}.}} \label{milestone}\vspace{-0.4cm}
\end{figure*}

A brief demonstration of the OWOD problem is shown in Fig. \ref{Demo}. The subsequent OWOD methods \cite{OW-DETR, PROB,2B-OCD,DOWB,OW-RCNN,RandBox} were proposed and gradually formalized the benchmark of this area. The 80 classes of the MS-COCO dataset are split into four tasks distinct by the semantic super-category, and data from the classes of each task is selected as the training dataset. 20 categories are introduced to the model for each task. Unknown recall and mean average precision are reported as the results for novel and previously known categories.

\wq{To better understand the development and foundation of OWOD, it's important to review its origins in OSR and ILOD. A chronological list of milestones of such area is concluded in Fig. \ref{milestone}. The foundational concepts for OWOD are built on prior works in OSR and ILOD. Scheirer \textit{et al.} pioneered OSR in 2012 with Towards OSR \cite{OSR}, followed by deep learning-based methods like OpenMax \cite{OSDN} and CROSR \cite{CROSR}. ILOD \cite{ILOD}, introduced in 2017, was enhanced by Faster ILOD \cite{FasterILOD} for improved speed and accuracy. Other methods such as RODEO \cite{acharya2020rodeo} and iOD-ML \cite{joseph2021incremental} have shown better performance. In 2021, Joseph \textit{et al.} \cite{ORE} synthesized OSR and ILOD concepts to propose OWOD, defining a model to detect both known and unknown classes and incrementally learn the identified unknown ones, inspired by the dynamic nature of real-world knowledge growth. This synthesis marked a pivotal milestone, leading to the development of various OWOD methodologies such as OW-DETR \cite{OW-DETR}, RE-OWOD \cite{RE-OWOD}, UC-OWOD \cite{UC-OWOD}, PROB \cite{PROB}, CAT \cite{CAT}, OCPL \cite{OCPL}, and OW-RCNN \cite{OW-RCNN}.}\label{sec1}

\wq{Although many comprehensive review papers on conventional object detection exist, such as \cite{ODreview1, ODReview2, CODReview, SalientReview, wu2024towards}, they do not fully address OWOD problems. Related surveys like those by Geng \textit{et al.} \cite{OSRreview} and Tian \textit{et al.} \cite{FSCILReview} cover open set recognition and few-shot class-incremental learning, respectively, but do not reflect the incremental learning aspect of OWOD. Boult \textit{et al.} \cite{OWRReview} extended OSR to Open World Recognition, including most open-set deep networks. Wu \textit{et al.} \cite{wu2024towards} discuss open vocabulary object detection, which emphasizes leveraging text information and word embeddings of unknown classes, whereas OWOD does not have this additional information, thus focusing on a system's ability to adapt to new categories in dynamic environments. Furthermore,}
as the number of related methods increases, 
\wq{various} data splits and evaluation metrics \wq{have been} proposed. \wq{Consequently}, there is no comprehensive survey paper \wq{that summarizes}  all Open World Object Detection methods. 



\begin{figure}
    \centering
    \includegraphics[scale=0.3]{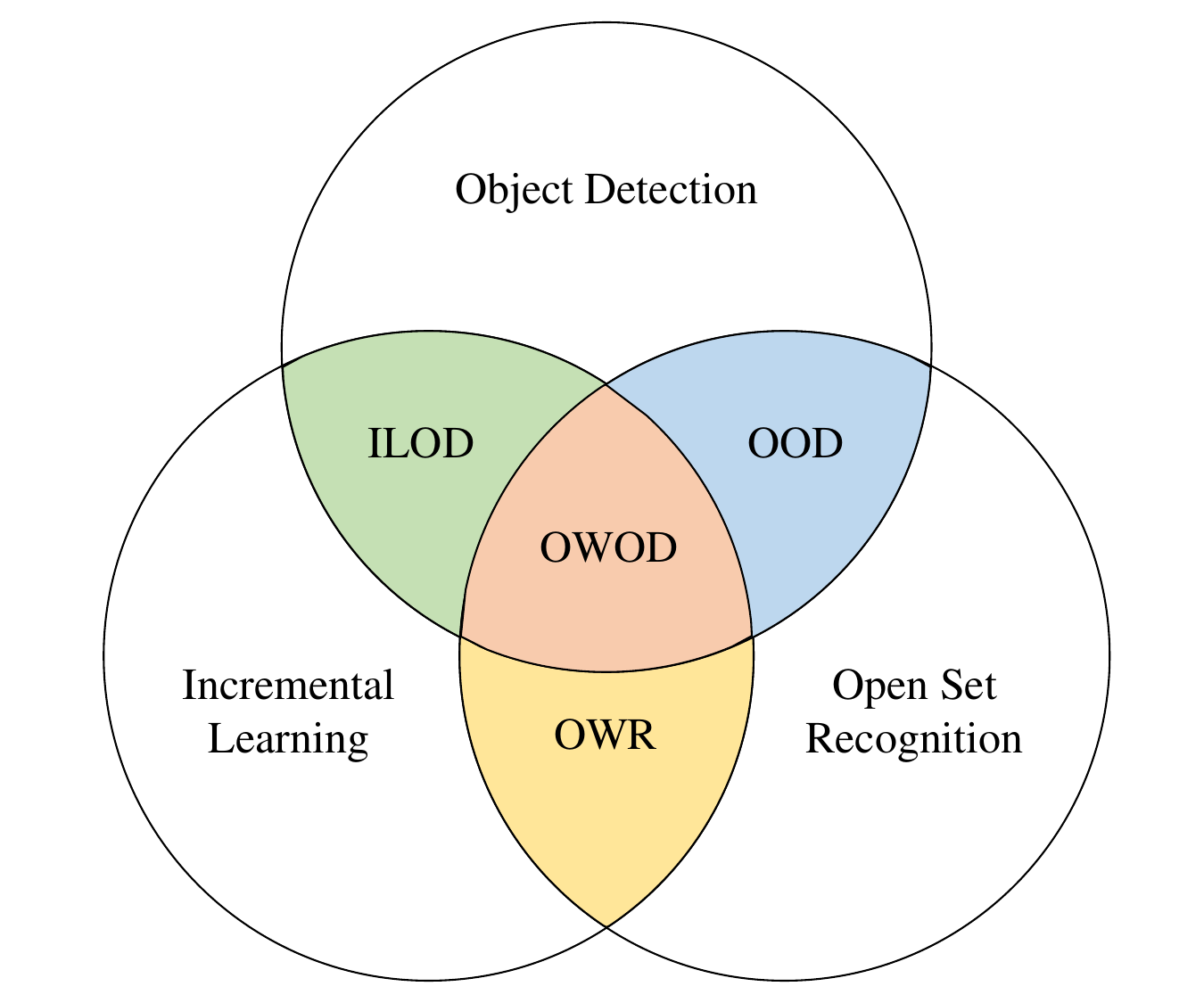}
    \caption{Relationship of related areas. \ym{Open World Object Detection (OWOD) combines the features of Object Detection, Incremental Learning, and Open Set Recognition. These three research areas are combined in pairs into Incremental Object Detection (ILOD), Out-of-distribution Detection (OOD), and Open World Recognition (OWR) research topics.}}
    \label{fig:venn1}
\end{figure}


\ym{This survey paper is intended for researchers and practitioners in the fields of computer vision and machine learning, particularly those interested in OWOD. The scope of this paper covers the fundamental concepts, core challenges, mainstream methods, benchmarks, and future research directions in OWOD. By systematically reviewing and summarizing existing OWOD works, this survey aims to provide a comprehensive reference for relevant researchers and practitioners, helping them quickly grasp the current research status and development trends in OWOD and offering insights and inspiration for future research in this area.}
Compared to previous review papers \cite{OSRreview, FSCILReview}, this paper is the first to review the field of OWOD, covering the latest models and methods in the OWOD area. We adopt a perspective different from that of other literature, i.e.\cite{OSRreview, FSCILReview, ODReview2}, we combine OSR with ILOD while considering the characteristics of open and incremental learning. By thoroughly analyzing and summarizing existing papers in the OWOD direction, we propose a novel taxonomy method for OWOD algorithms according to the techniques used on region proposal and unknown-aware classification modules. Specifically, we first introduce the backbone networks, baselines, and related areas such as incremental learning and open set recognition in Section \ref{sec2}. In Section \ref{sec3}, we present a comprehensive review of existing approaches to Open World Object Detection, including their strengths and limitations. Standard benchmarks, including datasets and evaluation metrics, are described in Section \ref{sec4}. Based on the benchmarks, a comprehensive comparison of state-of-the-art methods is shown also in Section \ref{sec4}. Finally, we discuss current challenges and future directions in this area and the potential impact of this work in Section \ref{sec5}, and we conclude this paper in Section \ref{sec6}.



\wq{The main contributions of this review are summarized as follows:}

\begin{itemize}
\item \wq{To the best of our knowledge, this is the first survey paper that comprehensively reviews the OWOD methods, addressing the urgent need to investigate, compare, analyze, and summarize the current state of research in this area.}
\item \wq{We propose a novel taxonomy method for OWOD algorithms, categorizing techniques based on the techniques adopted in the region proposal and unknown-aware classification modules.}
\item \wq{Our paper uniquely combines concepts from Open Set Recognition (OSR), Incremental Learning Object Detection (ILOD), and Open Vocabulary Object Detection (OVOD), integrating the characteristics of open and incremental learning to offer a new perspective on OWOD.}
\item \wq{We describe standard benchmarks, including datasets and evaluation metrics, and provide a comprehensive comparison of state-of-the-art OWOD methods based on these benchmarks.}
\end{itemize}
\section{Related Work}\label{sec2}
In this section, we review the related works of Open World Object Detection (OWOD), including the commonly used backbone networks in Section \ref{sec2a}, object detection baselines in Section \ref{sec2b}, open set recognition in Section \ref{sec2c}, and incremental learning in Section \ref{sec2d}. A Venn graph indicating the relations among the related areas of OWOD is illustrated as Fig.~\ref{fig:venn1}.

\subsection{Backbone Networks}\label{sec2a}
A “backbone” refers to a deep neural network architecture that processes input images and extracts features. Many neural network backbones are currently publicly available and open-source. Several backbones are overwhelming due to their outstanding performance and efficiency, such as Very Deep Convolutional Networks from Visual Geometry Group (VGGNet)\cite{simonyan2015very}, Residual Network (ResNet)\cite{he2016deep}, MobileNet\cite{howard2017mobilenets}, and EfficientNet\cite{tan2019efficientnet}.

\bb{The choice of backbone network can significantly impact the performance of object detection models. In the context of OWOD, the backbone network should be able to capture discriminative features for both known and unknown classes. Recent advancements in backbone architectures, such as the introduction of Transformer-based models like ViT~\cite{dosovitskiy2021image} and DETR~\cite{DETR}, have shown promising results in object detection tasks. These models leverage self-attention mechanisms to capture long-range dependencies and global context, which can be particularly beneficial for detecting objects in complex scenes. The evolution of backbone networks, from early architectures like VGGNet to more advanced models like ResNet and Transformers, has been instrumental in improving object detection and recognition performance.
}

\subsection{Object Detection Baselines}
\label{sec2b}
 \bb{The development of deep learning-based object detection has significantly evolved over the years, and object detection methods are surveyed comprehensively over several works such as \cite{ODreview1,ODReview2,CODReview,OSRreview,yilmaz2006object,ragland2014survey,liu2020deep}. Initially, object detection models relied on hand-crafted feature extractors like the Viola-Jones detector \cite{viola2001rapid} and Histogram of Oriented Gradients (HOG)\cite{dalal2005histograms}, which were slow and inaccurate. The advent of Convolutional Neural Networks (CNNs) and their application in image classification, especially with the success of AlexNet \cite{krizhevsky2012imagenet} in the ImageNet Large Scale Visual Recognition Challenge (ILSVRC) 2012, marked a turning point in visual perception. This led to further research and development in computer vision, resulting in a variety of object detection applications, including self-driving, facial detection, and security systems.}
 
 \bb{Over time, object detection models have evolved into single-stage and two-stage detectors, each with its advantages and limitations. Single-stage detectors like You Only Look Once (YOLO) \cite{redmon2016yolo9000} are known for their speed and suitability for real-time applications. Meanwhile, two-stage detectors, such as Faster R-CNN\cite{FasterRCNN}, tend to be more accurate but slower. Modern advancements have also focused on developing lightweight models for deployment on mobile and embedded systems, highlighting the need for efficient and scalable solutions. For instance, EfficientDet\cite{tan2020efficientdet} achieves high accuracy and efficiency with its scalable design. Feature Pyramid Network (FPN) and its variation \cite{lin2017feature,gong2021effective} provide scalability for the detection of different sizes of objects, which promotes the development of other object detection applications \cite{yu20201st}. Swin Transformer \cite{liu2021swin} introduces transformer-based backbones that promise a paradigm shift from traditional CNNs.} Among these tremendous works, simple yet good-performing object detection models (or baseline models) are developed upon these backbone architectures, providing a reference for evaluating the effectiveness of new techniques and architectures, including OWOD. Thus, we review various methods using different backbones and baselines.

Region-based Convolutional Neural Network (R-CNN)\cite{girshick2014rich} identifies and locates objects in images. It operates in two stages: region proposal generation and object classification. First, potential regions of interest (RoI) which contain objects are proposed. Then the classification branch will classify these regions to a specific object category accordingly. 

R-CNN's challenge is its slow inference due to using selective search for proposals. Fast R-CNN\cite{girshick2015fast} introduced RoI pooling, which shares convolutional features across proposals. This shared approach speeds up detection by reducing redundancy. Faster R-CNN\cite{FasterRCNN} brought in a Region Proposal Network (RPN), a subnetwork generating proposals directly, making the architecture end-to-end trainable.

Mask-RCNN\cite{he2017mask} added a branch to predict object masks alongside Faster R-CNN. This mask branch delineates object boundaries, enhancing R-CNN's segmentation capabilities.

Faster R-CNN\cite{FasterRCNN} can integrate various feature extractors like VGG and ResNet. Other R-CNN derivatives include Feature Pyramid Network (FPN)\cite{lin2017feature}, which blends multi-scale feature maps; Cascade R-CNN\cite{cai2018cascade}, refining detection with rising IoU thresholds; and Selective Nested Iterative Pooling (SNIP) algorithm \cite{singh2018sniper}, which selectively pools region proposals to detail objects at multiple scales.

Transformer\cite{vaswani2017attention} based methods have been widely used in all learning-based studies due to the high potential of the self-attention mechanism. It is also widely accepted and studied in the field of OWOD. 
The Detection Transformer (DETR)\cite{DETR} is a transformer-based model that directly predicts object bounding boxes and class labels without relying on components like anchor boxes. DETR uses a backbone CNN to extract features from the input image, which are then processed by a transformer encoder-decoder architecture. Self-attention mechanisms enable the model to capture relationships between objects and global context, improving localization and recognition accuracy. Building upon DETR, Deformable DETR\cite{DDETR} incorporates deformable attention modules. These modules enhance the capability of the transformer to model spatial relationships by allowing adaptive sampling locations within the input feature map. Deformable DETR improves localization accuracy by addressing the limitations of the regular grid-based attention mechanism. One key design for DETR is the one-to-one set matching to establish its end-to-end capability so that object detection does not require a hand-crafted Non-Maximum Suppression (NMS) to remove duplicate detections. The state-of-the-art variant H-DETR\cite{jia2023detrs} imposes a hybrid matching scheme that combines the original one-to-one matching branch with an auxiliary one-to-many matching branch during training. SETR \cite{zheng2020rethinking} focuses on addressing the limitations of regular grid-based attention in object detection. It incorporates a position embedding module that encodes spatial coordinates of object proposals, enabling the model to reason about the spatial layout of objects. \bb{Co-DETR\cite{zong2023detrs} provides a new training frame utilizing multiple parallel auxiliary heads supervised by one-to-many label assignments, further boosting the efficiency and effectiveness of DETR-based methods.}

\subsection{Open Set Recognition}\label{sec2c}
As an extension of the Open Set Recognition (OSR) problem, Open World Object Detection (OWOD) inherits some features from OSR. Thus, an overview of OSR can help to get a better understanding of OWOD. Open set recognition is a scenario where the training data is incomplete in classes, and there are unknown classes that can be submitted to the algorithm during testing. This means that classifiers must be able to accurately classify known classes and effectively handle unknown classes. In simple terms, OSR is the ability to determine whether a test sample belongs to one of the classes that the classifier was trained on or not. Out-of-distribution (OOD) Detection, or Open Set Detection, is also introduced in this section.

As illustrated in \cite{OSR}, regions distant from recognized data, encompassing both known known classes (KKCs) and known unknown classes (KUCs), are typically identified as the open space, $\mathcal{O}$. Assigning any data point within this domain to a random KKC carries an inherent risk, termed the open space risk, $R_{\mathcal{O}}$. Since unknown unknown classes (UUCs) are not accounted for during training, gauging the open space risk quantitatively often poses challenges. Instead, \cite{OSR} offers a qualitative outline of $R_{\mathcal{O}}$. Here, $R_{\mathcal{O}}$ is defined in relation to the size of open space $\mathcal{O}$ versus the entire measurement space $S_{o}$:
\begin{equation}\label{OpenSpaceRisk}
    R_\mathcal{O}(f)=\frac{\int_{\mathcal{O}}^{}f(x)dx}{\int_{S_{o}}^{}f(x)dx}
\end{equation}
In this equation, $f$ represents a measurable identification function. When $f(x) = 1$, it suggests the detection of a class within KKCs, and if $f(x) = 0$, no class is recognized. Following this definition, a higher frequency of labeling data in the open space as KKCs leads to an increased value of $R_{\mathcal{O}}$. Assume $V$ represents the training dataset. Let $R_{\mathcal{O}}$ stand for the open space risk, while $R_{\varepsilon}$ symbolizes the empirical risk. In open set recognition, the objective is to identify a measurable recognition function, $f$, within the set $\mathcal{H}$, such that when $f(x)>0$, it indicates correct recognition. The function $f$ is derived by aiming to reduce the associated Open Set Risk. To formalize this problem, the open set risk is defined in \cite{OSR} as Eq. (\ref{OpenRisk}).
\begin{equation}\label{OpenRisk}
    \textup{arg}\underset{f\in \mathcal{H}}{\textup{min}}\left \{ R_{O}(f)+\lambda _{r}R_{\varepsilon }(f(V)) \right \}.
\end{equation}
where $\lambda _{r}$ is a regularization constant. Thus, the objective of the OSR problem is to identify a quantifiable recognition function that minimizes open set risk.

The OSR algorithms can be categorized into two classes according to \cite{OSRreview}: discriminative models and generative models. From the perspective of discriminative models, there are two subcategories: Traditional Machine Learning (TML)-based algorithms and Deep Neural Network (DNN)-based algorithms. For generative models, instance generation and non-instance generation-based algorithms are two subcategories. Most existing works focus on DNN-based OSR. A detailed summary of different OSR methods is presented as follows.

\begin{table}[htbp]
  \centering
  \vspace{-2mm}
  \caption{Taxonomy of Open Set Recognition methods}
  \vspace{-2mm}
    \begin{tabular}{m{0.1\textwidth}<{\centering}m{0.15\textwidth}<{\centering}l}
    \toprule
    \multicolumn{2}{c}{Categories of OSR methods} & \multicolumn{1}{c}{Papers} \\
    \midrule
    \multirow{2}{0.1\textwidth}{Discriminative Model} & Traditional ML-based & \multicolumn{1}{c}{\cite{OSR,W-SVM,SROSR,OWR,EVM}} \\
\cmidrule{2-3}          & Deep Neural Network-based & \multicolumn{1}{c}{\cite{OSDN,Hassen,OSRText,DOC,COOL,C2AE}} \\
    \midrule
    \multirow{2}{0.1\textwidth}{Generative Model} & Instance Generation-based & \multicolumn{1}{c}{\cite{G-OpenMax,ASG}} \\
\cmidrule{2-3}          & Non-Instance Generation-based & \multicolumn{1}{c}{\cite{CD-OSR}} \\
    \bottomrule
    \end{tabular}%
  \label{tab:OSR}%
\end{table}%

\subsubsection{TML-based Open Set Recognition}

In OSR scenarios, the assumption that training and testing data are drawn from the same distribution no longer holds. These methods aim to adapt traditional machine learning methods to OSR. These methods can be categorized into several groups based on the traditional machine learning method.

SVM-based methods adapt the Support Vector Machine (SVM) to the OSR scenario by incorporating an open space risk term in modeling to account for the space beyond the reasonable support of KKCs. 1-versus-Set machine \cite{OSR} incorporates an open space risk term in modeling by adding another hyperplane in parallel with the separating hyperplane obtained by SVM in score space. Weibull-calibrated SVM (W-SVM) \cite{W-SVM} implements non-linear kernels into a solution that further limited open space risk by positively labeling only sets with finite measure and combines the statistical extreme value theory (EVT) for score calibration with two separated SVMs. Sparse Representation-based methods utilize sparse representation-based techniques for OSR. Sparse representation-based open set recognition model (SROSR) \cite{SROSR} models the tails of the matched and sum of non-matched reconstruction error distributions using EVT. However, it has limitations, such as failing in cases where the dataset contains extreme variations in pose, illumination, or resolution. Distance-based methods attempt to implement distance-based classifiers in the OSR scenario. Nearest Non-Outlier (NNO) \cite{OWR} extends upon the Nearest Class Mean (NCM) classifier by carrying out classification based on the distance between the testing sample and the means of KKCs. It can dynamically add new classes based on manually labeled data. Margin Distribution-based methods utilize margin distributions to provide better error bounds than those offered by a soft-margin SVM. Extreme Value Machine (EVM) \cite{EVM} stems from the concept of margin distributions and obtains a theoretically sound classifier by extending margin distribution theory from a per-class formulation to a sample-wise formulation. 

\subsubsection{DNN-based Open Set Recognition}

With the development of more powerful computational resources, the deep neural network plays an important part in many application scenarios, including open set recognition. However, DNNs often make wrong predictions when processing unknown unknown classes (UUCs) samples due to their inherent closed-set nature. The following DNN-based OSR methods use different approaches to address this problem.

Bendale and Boult proposed OpenMax \cite{OSDN}, which replaces the SoftMax layer with an OpenMax layer and represents each class as a mean activation vector (MAV) with the mean of the activation vectors in the penultimate layer of that network. However, it fails to recognize adversarial images and has limitations, such as not directly incentivizing projecting class samples around the MAV. Hassen and Chan \cite{Hassen} proposed a neural network-based representation for open set recognition to address the inaccurate measurement due to the inconsistency of testing and training distance function. Following OpenMax, Prakhya \textit{et al.} \cite{OSRText} explored the open set text classification. Shu \textit{et al.} \cite{DOC} presented the Deep Open classifier (DOC) model by replacing the SoftMax layer with a 1-versus-rest final layer of sigmoids. Kardan and Stanley \cite{COOL} proposed a competitive overcomplete output layer (COOL) neural network, which circumvents the overgeneralization of neural networks over regions far from training data. C2AE model \cite{C2AE} proposed by Oza and Patel uses class-conditioned auto-encoders with novel training and testing methodology.

\subsubsection{Instance Generation-based Open Set Recognition}

Instance generation-based OSR methods aim to account for open space with UUCs generated by adversarial learning (AL) techniques. These methods employ a generative model and a discriminative model, where the generative model learns to generate samples that can fool the discriminative model as non-generated samples. Ge \textit{et al.} proposed the Generative OpenMax (G-OpenMax) \cite{G-OpenMax} to synthesize mixtures of UUCs using a conditional generative adversarial network (GAN) and provide explicit probability estimation over the generated UUCs. Adversarial Sample Generation (ASG) framework \cite{ASG} proposed by Yu \textit{et al.} can be applied to various models besides neural networks and can generate data of both UUCs and KKCs if necessary. 

\subsubsection{Non-Instance Generation-based Open Set Recognition}

As a non-instance generation-based OSR method, Dirichlet process-based OSR methods \cite{CD-OSR} aim to adapt the Dirichlet process (DP), a stochastic process that has been widely used in clustering and density estimation problems as a nonparametric prior defined over the number of mixture components, to the OSR scenario. Geng and Chen \cite{CD-OSR} proposed a collective decision-based OSR model (CD-OSR), which adapts the hierarchical Dirichlet process (HDP) to OSR and can address both batch and individual samples. In the training phase, CD-OSR conducts a co-clustering process to determine the appropriate parameters. During the testing phase, it models the data of each KKC as a group of CD-OSR using a Gaussian mixture model (GMM) with an unspecified number of components or subclasses. After co-clustering is completed, one or more subclasses representing the corresponding class can be identified. A testing sample is then classified as either the appropriate KKC or a UUC, depending on whether its assigned subclass is associated with the corresponding KKC.

\subsubsection{\bb{Out-of-distribution Detection}}
\ym{Out-of-distribution (OOD) detection methods can be broadly categorized into classification-based methods, density-based methods, distance-based methods, and reconstruction-based methods. Each category utilizes a different approach to identify samples that deviate from the training distribution. Classification-based methods include output-based techniques and outlier exposure. For instance, ODIN \cite{liang2017enhancing} uses temperature scaling and input perturbations to improve the separation between in-distribution and OOD samples, while LogitNorm \cite{wei2022mitigating} enforces a constant vector norm on logits during training to produce more reliable confidence scores. Outlier Exposure (OE) \cite{hendrycks2018deep} utilizes external OOD datasets during training to enhance detection capabilities. Density-based methods explicitly model the density of in-distribution data, flagging low-density regions as OOD. An example is the use of class-conditional Gaussian distributions \cite{lee2018simple}, which identify OOD samples based on their likelihoods within the modeled distribution. Distance-based methods calculate the distance between test samples and class prototypes or centroids in the feature space. Some examples include Mahalanobis distance-based OOD detection \cite{lee2018simple} and KNN-based methods \cite{sun2022out}, which use nearest neighbor distances without assuming any specific distribution for the feature space. Reconstruction-based methods rely on autoencoders or similar techniques to detect OOD samples by analyzing reconstruction errors, such as autoencoder-based methods \cite{chen2018autoencoder} and MoodCat \cite{yang2022out}.}

In conclusion, OSR achieves part of the goals of OWOD. Such ideas can be implemented for OWOD to obtain an accurate result of unknown detection and classification. However, the OSR or OOD models cannot update the knowledge when newly labeled UUCs are presented, which should be integrated with incremental learning in the following sections.
\subsection{Incremental Learning}\label{sec2d}
Incremental learning is a machine learning method that allows incremental updates to existing models without retraining the entire model. In OWOD, incremental learning can gradually improve the model's performance by receiving new data or new tasks without causing too much interference to existing knowledge. This approach is very useful in many practical applications, especially when the data is growing or the task is changing.

To gain new knowledge while keeping existing knowledge, incremental learning should overcome the stability-plasticity dilemma. There is a common flaw in model training, known as catastrophic forgetting, where machine learning models (especially deep learning methods based on backpropagation) typically exhibit significant performance degradation on previous tasks when trained on new tasks. One of the main reasons for catastrophic forgetting is that traditional models assume that data distribution is fixed or stationary and that training samples are independent and identically distributed. Therefore, the model can see the same data for all tasks repeatedly. However, when the data becomes a continuous data stream, the distribution of the training data is non-stationary. As the model continually learns from this non-stationary data distribution, new knowledge interferes with old knowledge, leading to a rapid decline in model performance and even the complete coverage or forgetting of previously learned knowledge. 

Incremental learning, depending on its algorithms, can be classified into three categories, i.e., regularization-based methods, replay-based methods, and parameter isolation methods. The paradigm of regularization-based and replay-based incremental learning has received more attention. The parameter isolation paradigm requires introducing more parameters and computational complexity and thus is typically used for simple task incremental learning.

\begin{table}[htbp]
  \centering
  \vspace{-2mm}
  \caption{Taxonomy of Incremental Learning methods}
  \vspace{-2mm}
    \begin{tabular}{ll}
    \toprule
    Incremental Learning Categories & \multicolumn{1}{c}{Papers} \\
    \midrule
    Regularization-based & \multicolumn{1}{c}{\cite{li2017learning, rannen2017encoder, kirkpatrick2017overcoming}} \\
    \midrule
    Replay-based & \multicolumn{1}{c}{\cite{rebuffi2017icarl,castro2018end,wu2019large,lopez2017gradient,acharya2020rodeo}} \\
    \midrule
    Knowledge distillation-based & \multicolumn{1}{c}{\cite{ILOD,FasterILOD,li2019rilod}} \\
    \midrule
    Few-shot / meta-learning-based & \multicolumn{1}{c}{\cite{perez2020incremental,joseph2021incremental}} \\
    \bottomrule
    \end{tabular}%
  \label{tab:IL}%
\end{table}%

\subsubsection{Regularization-based Incremental Learning}
The main idea of regularization-based incremental learning is to protect old knowledge from being overwritten by new knowledge, by applying constraints to the loss function of the new task. Such methods typically do not require the model to revisit previously learned tasks with old data. Learning without Forgetting (LwF)\cite{li2017learning} algorithm is a typical regularization-based method. This idea is derived from knowledge distillation, which makes the predictions of the new model on the new task similar to the predictions of the old model on the new task by a distillation loss. However, the drawback of this approach is that it heavily relies on the correlation between the old and new tasks, and inter-task confusion may occur when the differences between tasks are too significant. Some researchers have proposed various improvement strategies based on the LwF algorithm. Some well-known approaches include the Encoder Based Lifelong Learning (EBLL)\cite{rannen2017encoder} algorithm, which is based on low-dimensional feature mapping, and the Elastic Weight Consolidation (EWC)\cite{kirkpatrick2017overcoming} algorithm, which is based on a Bayesian framework. The EWC algorithm corresponds to a general parameter constraint method. In summary, the regularization-based incremental learning methods correct the gradient by introducing additional loss to protect the old knowledge learned by the model, providing a way to alleviate catastrophic forgetting under specific conditions. However, although current deep learning models are over-parameterized, model capacity is still limited, and we usually need to strike a balance between the performance of old and new tasks.

\subsubsection{Replay-based Incremental Learning}
The basic idea of replay-based incremental learning is to review the old data. When training a new task, a representative subset of old data is retained and used to review the old knowledge learned by the model. Therefore, which part of the old task data to retain and how to train the model using both old and new data is the main issue that these methods need to consider. iCaRL\cite{rebuffi2017icarl} is the most classic replay-based incremental learning model, and its idea is similar to LwF. It also introduces distillation loss to update model parameters but relaxes the constraint of completely not using old data. Some\cite{castro2018end}\cite{wu2019large} algorithms dynamically adjust the number of preserved old data to avoid the linear growth of computational cost with the increase of task number, thus avoiding the drawback of linearly increasing computational cost in LwF algorithm. The incremental learning method of iCaRL updates the parameters of old tasks, which may lead to overfitting the retained old data by the model. GEM (Gradient Episodic Memory)\cite{lopez2017gradient} is then proposed to update the parameters only for the new task without interfering with the parameters for the old tasks. GEM uses inequality constraints to modify the gradient update direction for the new task, hoping the model can minimize the new task loss without increasing the loss for the old tasks. In general, the main disadvantage of replay-based incremental learning is that it requires additional computational resources and storage space to recall old knowledge. When the number of tasks continues to increase, either the training cost will increase, or the weights of representative samples will be weakened. 

Overall, the advantage of incremental learning is that new data can be trained at any time without retaining a large amount of training data, which makes storage and computation costs relatively low. Additionally, it can effectively avoid privacy leakage issues, which is very valuable in the context of edge computing. However, current incremental learning is still a very open research problem and is largely in the theoretical exploration stage.

\subsubsection{Incremental Learning Object Detection}
Incremental learning can also be used in the object detection domain to deal with open world object detection problems. Traditional object detection models are usually trained from scratch on a fixed dataset, but incremental learning allows the model to adapt to new information without discarding previously learned knowledge.

Distillation-based methods are widely used for catastrophic forgetting. ILOD\cite{ILOD} is first proposed to deal with this forgetting problem. It contains a frozen copy of the original detector to select proposals corresponding to the old classes and compute the distillation loss. It also contains a new adapted network for the new classes. To avoid catastrophic forgetting, two networks are connected by the proposed biased distillation. To accelerate the process on edge, RILOD\cite{li2019rilod}  proposed a bounding box distillation method with a constraint between old and new models. Learning a new object class can finish in less than 2 minutes on a single GPU with superior detection accuracy. Similarly, Faster ILOD\cite{FasterILOD} proposed an adaptive distillation with several internal connections.

A replay-based method\cite{acharya2020rodeo} is also proposed for object detection. The compressed images will be stored in the buffer to guide training later. And the feature extractor part of the model will be frozen after the early-stage training. Thus, only the classifier part of the network will be trainable for incremental learning.

Due to the long tail effect, most new classes of objects come with a limited quantity. Many research also consider few-shot learning or meta-learning-based methods to solve this problem. \cite{perez2020incremental} A few-shot learning method is proposed to incrementally recognize novel classes using only a few labeled examples per class. CentreNet is then proposed to reformulate object detection as a point+attribute regression problem. The key merit of CentreNet is that each individual class maintains its own prediction heatmap and makes independent detection by activation thresholding. To enroll new classes, it builds a meta-learning-based network to generate object-specific weights from the support set (few-shot), and the object locator uses these to detect objects in test images. Similar work \cite{joseph2021incremental} uses meta-learning and sets some of the layers as wrap layers. These wrap layers have better generalization to new tasks, faster convergence, and alleviating catastrophic forgetting. 
\subsection{\ym{Open Vocabulary Object Detection}}\label{sec2e}
Open Vocabulary Object Detection (OVOD) aims to enhance object detection models to recognize both known and novel object classes without predefined labels. These methods are categorized into five primary areas: knowledge distillation, region text pre-training, training with more balanced data, prompting modeling, and region text alignment according to \cite{wu2024towards}. Each category deploys different strategies to use the large-scale knowledge embedded in Vision Language Models (VLMs) to augment the capabilities of traditional, close-set object detectors.

\subsubsection{Knowledge Distillation}
This category leverages extensive knowledge embedded in VLMs to enhance close-set detectors to identify novel object categories. For example, the ViLD method\cite{gu2021open} combines a dual-branch mechanism involving text and image branches to facilitate visual-to-visual knowledge transfer. Methods such as HierKD\cite{ma2022open} and LP-OVOD \cite{pham2024lp} further refine this process by introducing loss modifications and extending the framework to incorporate new modules such as pseudo-labeling and global-level distillation modules. These modifications aim to bridge the gap between high-capacity VLMs and close-set detection models, enhancing the generalization capabilities of object detection models across different visual domains.

\subsubsection{Region Text Pre-training}
This strategy exploits a large number of available image-text pairs and converts them into rich training resources for object detection models. By aligning text and image features at the region level, models such as OVR-CNN\cite{zareian2021open} and attribute-sensitive OVR-CNN\cite{buettner2023enhancing} learn to map these features into a shared semantic space, thereby significantly improving the detection of new categories. This category also includes other methods such as GLIP\cite{li2022grounded, zhang2022glipv2} which utilize self-training techniques to generate ground-truth, thereby enhancing detection and grounding capabilities through extensive pre-training on large-scale datasets. Additionally, RO-ViT\cite{kim2023region} introduces a pretraining method that randomizes cropping and resizing regions of positional embeddings, and replaces the common softmax cross-entropy loss with focal loss.

\subsubsection{Training with More Balanced Data}
To address the challenge of data imbalance in training datasets, this category focuses on strategies to enhance model performance for common and rare object categories. Methods like Detic\cite{zhou2022detecting} use image-level supervision to better utilize object-centric classification data, while MM-OVOD\cite{kaul2023multi} introduces multi-modal text embeddings as classifiers to enrich the feature extraction process. Furthermore, some methods generate pseudo bounding box annotations from large-scale image-caption pairs (e.g., PB-OVD\cite{gao2022open}), leveraging advanced activation mapping techniques to improve accuracy on training data.

\subsubsection{Prompting Modeling}
This innovative approach adapts base models to specific tasks by incorporating cues that effectively guide model focus. Methods such as PromptDet\cite{feng2022promptdet} and CORA\cite{wu2023cora} attempt to use different prompt structures and embedding strategies to improve how models interact with novel class descriptions. This approach focuses on enhancing the model’s ability to incorporate learned prompts into the base model, allowing the model to transfer its knowledge to downstream tasks more easily. The work by Du \textit{et al.} \cite{du2022learning} on detection prompt (DetPro) introduces a novel method to learn continuous prompt representations, incorporating a background interpretation scheme and context grading scheme to enhance detection performance.

\subsubsection{Region Text Alignment}
This category aims to enable fine-grained recognition capabilities, focusing on carefully aligning textual features with corresponding visual areas. OV-DETR \cite{zang2022open} introduces a transformer-based detection strategy that enhances alignment through an innovative matching mechanism. Methods such as DetCLIPv2 \cite{yao2023detclipv2} and F-VLM \cite{kuo2022f} build on this foundation, utilizing alignment strategies that leverage ensemble-based and individual region-to-text matching methods to improve the accuracy and robustness of object detection across different visual scenes.

When comparing OVOD to Open World Object Detection (OWOD), significant differences and similarities are evident. Both applications extend the capabilities of object detection systems, allowing them to operate in more dynamic and unpredictable environments. However, OVOD mainly uses open vocabulary, leveraging language and visual pre-training to recognize new objects. In contrast, OWOD not only detects known and unknown objects, but also gradually learns these unknown objects into novel known classes without forgetting previously learned classes. Class-agnostic region proposals, unknown-aware classification, and incremental learning of unknown classes are integrated to continuously update and adapt the detection model. While OVOD aims at broadening the spectrum of detectable objects through language and vision integration, OWOD addresses the additional challenge of incrementally adapting to new object classes as they become known, tackling issues of catastrophic forgetting and adaptation more explicitly.
\section{OWOD Methods}\label{sec3}

\begin{table*}
  \centering
  \caption{Feature summary of OWOD methods}
  \label{tab:Architecture}%
  \vspace{-2mm}
  \resizebox{\textwidth}{!}{
    \begin{tabular}{lccccc}
    \toprule
     Name & Detector & Venue & \multicolumn{1}{c}{Evaluation Metrics} & \multicolumn{1}{p{9.89em}}{Class-agnostic Region Proposal Category} & \multicolumn{1}{p{10.055em}}{Unknown-aware Classification Category} \\
    \midrule
     ORE\cite{ORE} & Faster R-CNN & CVPR 2021 & mAP/WI/A-OSE & Pseudo-labeling & Metric-learning \\
     OW-DETR\cite{OW-DETR} & Deformable DETR & CVPR 2022 & mAP/UR & Pseudo-labeling & Classification head \\
     Fast-OWDETR\cite{chen2022fast} & Deformable DETR & -     & mAP/UR & Pseudo-labeling & Classification head \\
     Open World DETR\cite{OpenWorldDETR} & Deformable DETR & arXiv 2022 & mAP/UR/WI/A-OSE & Pseudo-labeling & Classification head \\
     CAT\cite{CAT} & Deformable DETR & CVPR 2023 & mAP/UR/WI/A-OSE & Pseudo-labeling & Classification head \\
     RE-OWOD\cite{RE-OWOD} & Faster R-CNN & TCSVT 2023 & mAP/UR/WI/UDR/UDP & Pseudo-labeling & Metric-learning \\
     OCPL\cite{OCPL} & Faster R-CNN & ICIP 2022 & mAP/UR/WI/A-OSE & Pseudo-labeling & Metric-learning \\
     UC-OWOD\cite{UC-OWOD} & Faster R-CNN & ECCV 2022 & mAP/UR/WI/A-OSE & Pseudo-labeling & Metric-learning \\
     2B-OCD\cite{2B-OCD} & Faster R-CNN & HCMA 2022 & mAP/UR/WI/UDR/UDP & Class-agnostic & Classification head \\
     PROB\cite{PROB} & Deformable DETR & CVPR 2023 & mAP/UR/WI/A-OSE & Class-agnostic & Classification head \\
     OW-RCNN\cite{OW-RCNN} & Faster R-CNN & arXiv 2023 & mAP/UR/WI/A-OSE & Class-agnostic & Classification head \\
     OLN\cite{OLN} & Faster R-CNN & ICRA 2022 & AP/AR/AUC & Class-agnostic & - \\
     RandBox\cite{RandBox} & Faster R-CNN & ICCV 2023 & mAP/UR/WI/A-OSE & Class-agnostic & Classification head \\
     DOWB\cite{DOWB} & Deformable DETR & arXiv 2023 & mAP/UR/WI/A-OSE & Others & Classification head \\
     MAVL\cite{MAVL} & Deformable DETR & ECCV 2022 & mAP/UR & Others & Others \\
     STUD\cite{STUD} & Faster R-CNN & CVPR 2022 & mAP   & Others & Others \\
     UnSniffer \cite{liang2023unknown} & Detectron2 & CVPR 2023 & mAP/UR/WI/A-OSE/U-AP/U-F1 & Pseudo-labeling & Others \\
     Ma et al.\cite{ma2023annealing} & Faster R-CNN \& DETR & CVPR 2024 & mAP/UR/WI/A-OSE/UDR/UDP & Others & Metric-learning \\
    \bottomrule
    \end{tabular}%
  }
\end{table*}%

\begin{table}[htbp]
  \centering
  \scriptsize
  \begin{threeparttable}
  \caption{Categories of OWOD methods}
  \label{tab:taxonomy}%
  \vspace{-2mm}
    \begin{tabular}{m{0.07\textwidth}<{\centering}m{0.07\textwidth}<{\centering}m{0.15\textwidth}<{}m{0.11\textwidth}<{}}
    \toprule
          &       & \multicolumn{2}{c}{Class-agnostic Region Proposal} \\
\cmidrule{3-4}          &       & Pseudo labeling & Class-agnostic \\
\cmidrule{1-4}
    \multicolumn{1}{c}{\multirow{8}[3]{0.07\textwidth}{Unknown-aware Classification}} & \multicolumn{1}{c}{\multirow{4}[1]{0.07\textwidth}{Classification head}}& \cellcolor[rgb]{ .85,  .95,  .815}OW-DETR \cite{OW-DETR} & \cellcolor[rgb]{ .98,  .89,  .84}OW-RCNN \cite{OW-RCNN} \\
          &       & \cellcolor[rgb]{ .85,  .95,  .815}CAT \cite{CAT}  & \cellcolor[rgb]{ .98,  .89,  .84}2B-OCD\cite{2B-OCD} \\
          &       & \cellcolor[rgb]{ .85,  .95,  .815}Fast-OWDETR\cite{chen2022fast} & \cellcolor[rgb]{ .98,  .89,  .84}RandBox\cite{RandBox} \\
          &       & \cellcolor[rgb]{ .85,  .95,  .815}Open World DETR\cite{OpenWorldDETR} & \cellcolor[rgb]{ .98,  .89,  .84}PROB \cite{PROB} \\
\cmidrule{2-4}          & \multicolumn{1}{c}{\multirow{3}[2]{0.07\textwidth}{Metric Learning}} & \cellcolor[rgb]{ .79,  .93,  .98}ORE \cite{ORE}  \\
          &       & \cellcolor[rgb]{ .79,  .93,  .98}RE-OWOD\cite{RE-OWOD} &  \multicolumn{1}{c}{\multirow{1}[2]{0.07\textwidth}{N.A.}} \\
          &       & \cellcolor[rgb]{ .79,  .93,  .98}OCPL\cite{OCPL}  &  \\
          &       & \cellcolor[rgb]{ .79,  .93,  .98}UC-OWOD\cite{UC-OWOD} &  \\
    \bottomrule
    \end{tabular}%
    \begin{tablenotes}
        \item \colorbox[rgb]{ .85,  .95,  .815}{Green} indicates pseudo-labeling category, \colorbox[rgb]{ .98,  .89,  .84}{red} indicates class agnostic category and \colorbox[rgb]{ .79,  .93,  .98}{blue} includes metric-learning methods. Methods that use other techniques are categorized as Other. (Best viewed in color)
    \end{tablenotes}
    \end{threeparttable}
  \vspace{-0.4cm}
\end{table}%

Open World Object Detection (OWOD) is composed of three main tasks: class-agnostic region proposal, unknown-aware classification and unknown-class incremental learning. The class-agnostic region proposal are derived from open set recognition, and the unknown-class incremental learning is the open world version of class incremental object detection. In the first stage, different techniques are used to extract all objects from the background, regardless of their categories. During the unknown-aware classification stage, the previous known classes and unknown objects should be classified. In the unknown-class incremental learning stage, the ground truth labels of detected unknown objects will be provided, and such unknown classes will be learned as new known classes. Different methods will be performed to alleviate the catastrophic forgetting problem of previously learned classes.

In this section, we formalize the definition of the OWOD problem and its relations with open set recognition and class incremental learning. We perform a review of most OWOD methods and categorize them into different branches according to their method of unknown detection. In Table \ref{tab:Architecture}, we review the different features of most OWOD methods, including backbone, categories, venue, evaluation metrics, unknown classification, and unknown proposal categories. Most of the methods use an object detector baseline such as Faster R-CNN \cite{FasterRCNN} or Deformable DETR \cite{DDETR} as the backbone to extract the features of objects as listed. 
As most of the OWOD methods follow the training and evaluation protocol of ORE\cite{ORE}, the evaluation metrics they used are listed. For the categorization of different methods, two different processes, i.e., unknown proposal and unknown classification, are classified separately. 

\subsection{Problem Formulation}
We first define a situation of classical object detection. The set of object classes is $\mathcal{K}^{t}=\left \{ 1,2,...,\textup{C} \right \}\subset \mathbb{N}^{+}$, where $\mathbb{N}^{+}$ is a positive integer set. Given a specific time $t$, a model $\mathcal{M}_{C}$ is trained on a dataset $\mathcal{D}^{t}=\left \{ \mathcal{I}^{t}, \mathcal{L}^{t} \right \}$, where $\mathcal{I}^{t}=\left \{ I_{1},I_{2},I_{3},...,I_{N} \right \}$ denotes N input images and $\mathcal{L}^{t}=\left \{ L_{1},L_{2},L_{3},...,L_{N} \right \}$ denotes the corresponding N labels. For each label $L_{i}=\left \{ l_{1},l_{2},l_{3},...,l_{K} \right \}$, there are $K$ object instances, and each instance is composed of category label and bounding box locations, i.e., $l_{k}=[c_{k},x_{k},y_{k},w_{k},h_{k}]$, where $c_{k}\in \mathcal{K}^{t}$ is a one-hot vector for object classes. $x_{k},y_{k}$ is the coordinates of the bounding box center point, and $w_{k},h_{k}$ denotes the width and height of the bounding box, respectively.

Following the setup of ORE \cite{ORE} OWOD describes another scenario: There exists an unknown category set $\mathcal{U}=\left \{ \textup{C}+1,\textup{C}+2,...\right \}$. An OWOD model $\mathcal{M}^{t}$ is trained by dataset $\mathcal{D}^{t}$ and is able to detect all previously known $\textup{C}$ classes. Apart from known objects, the model might encounter other classes that belong to $\mathcal{U}$ during the test process. Thus, the model is trained to identify a novel or unseen instance by labeling the instance as an unknown or zero (0) category. This is the unknown proposal and unknown classification stage of Open World Object Detection. 

\begin{figure*}
\centering
\includegraphics[scale=0.45]{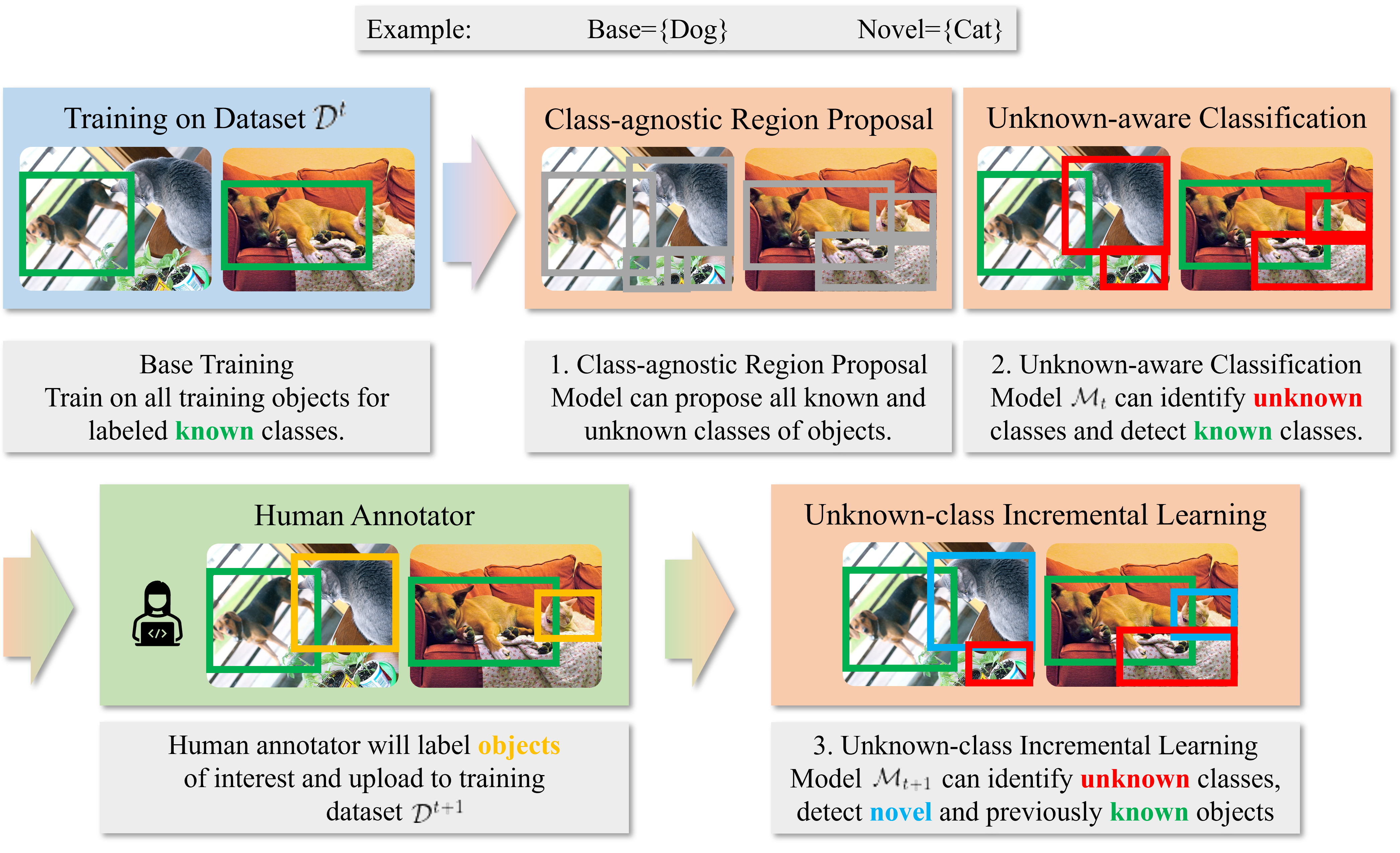}\vspace{-0.2cm}\caption{Framework of Open World Object Detection.} 
\label{Framework}
\vspace{-4mm}
\end{figure*}

In the unknown-class incremental learning stage, the previously detected set of unknown instances $U^{t}\subset \mathcal{U}$ will be sent to an oracle (e.g., a human annotator). The oracle will label $n$ novel classes of interest and generate a set of corresponding training samples. By incrementally adding labeled novel classes to known categories, the new set of known classes will be described as $\mathcal{K}^{t+1}=\mathcal{K}^{t}+\left \{ \textup{C}+1, ...,\textup{C}+n \right \}$, and the labeled novel objects will be used to form a new dataset $\mathcal{D}^{t+1}$. In the case of limited computing and memory resources in real-world scenarios, only a few training examples can be used for the previously known classes $\mathcal{K}^{t}$. After training by the labeled novel objects and few-shot previously known objects, not training from scratch by the whole dataset, the model $\mathcal{M}^{t}$ will be updated as $\mathcal{M}^{t+1}$, which is able to detect objects from new categories while alleviating catastrophic forgetting of previously known classes $\mathcal{K}^{t}$. This cycle can be repeated whenever new objects are encountered. The overall framework of the OWOD problem is illustrated in Figure \ref{Framework}. 


We examine OWOD methods currently available at our best in the literature. According to their methods of detecting the unknown object, we divide these methods into four categories: pseudo-labeling-based methods, class-agnostic methods, metric learning-based methods, and other methods. The taxonomy of each method is summarized in Table \ref{tab:taxonomy}. We present all categories of OWOD in detail as follows.

\begin{figure*}[ht]
\centering
\includegraphics[scale=0.4]{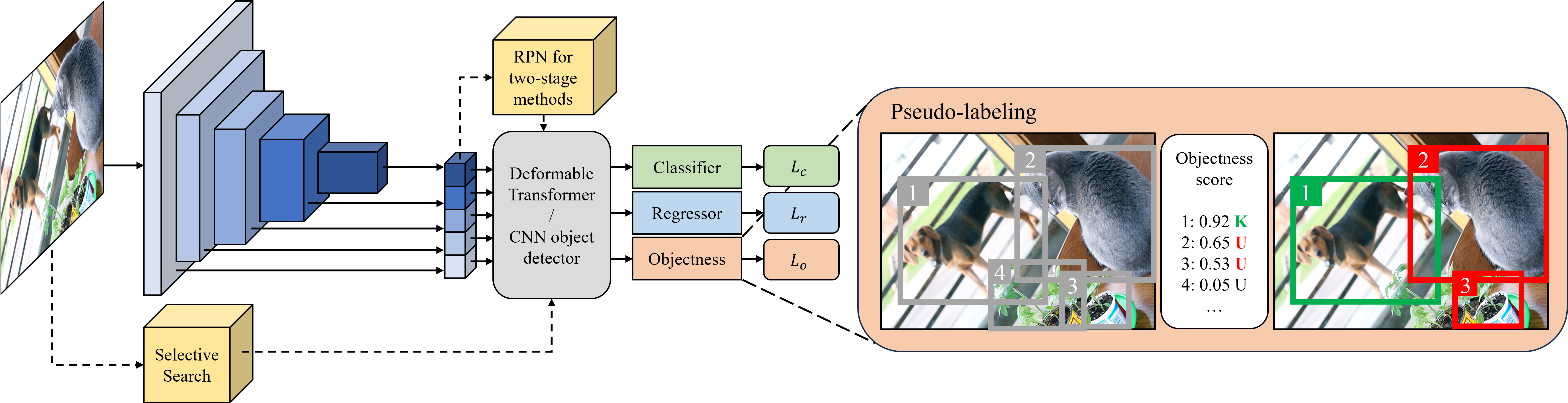}\caption{Framework of Pseudo-labeling-based OWOD methods. After feature extraction and bounding box proposal, a pseudo-labeling technique is used to provide labels for unknown objects. The selection of pseudo-labels is based on a self-defined objectness score. The yellow cubes with dashed arrows indicate the structures that might be adopted in various models. \textcolor[RGB]{0,176,80}{\textbf{K}} indicates that the proposal belongs to previously known category and \textcolor[RGB]{255,0,0}{\textbf{U}} indicates proposal from unknown category. RPN stands for Region Proposal Network. } \label{PL}\vspace{-0.4cm}
\end{figure*}

\subsection{Pseudo-labeling methods}\label{sec3b}

Pseudo-labeling-based methods adopt the pseudo-labeling technique to select unknown objects during the training process. They usually use a self-defined objectness score to measure whether the selected region contains an object or not. Object proposals with the top-k objectness scores and that do not match with known categories will be pseudo-labeled as unknown objects. A common structure of pseudo-labeling-based OWOD methods is shown in Fig. \ref{PL}. 


\ym{ORE \cite{ORE} is the first paper that proposes the OWOD problem, and also the first to use pseudo-labeling technique in OWOD area. Joseph \textit{et al.} proposed an auto-labeling scheme to label the background object proposals generated by the region proposal network with the top objectness scores but do not overlap with known ground truth as unknown objects. Using a two-stage Faster R-CNN \cite{FasterRCNN} as the backbone object detector, this paper introduces a strong evaluation protocol and provides a novel OWOD solution based on contrastive clustering and energy-based unknown identification. However, it is categorized as a metric learning-based method according to the clustering-based unknown-aware classification methods. The details of ORE will be presented in metric-learning-based methods.}

Followed by ORE \cite{ORE}, Gupta \textit{et al.} \cite{OW-DETR} proposed another pseudo-labeling-based OWOD method using Deformable DETR \cite{DDETR} as the backbone called Open World Detection Transformer \cite{OW-DETR} (OW-DETR). The authors argue that a single-stage transformer will introduce fewer inductive biases and can encode long-range dependencies at multi-scales. Also, no supervision for unknown instances makes it closer to real open-world settings. To achieve this, OW-DETR first deploys Deformable DETR as multi-scale context encoding to encode richer context over a larger receptive field. Second, a bottom-up attention-driven pseudo-labeling scheme is deployed to better detect unknown classes. The top queries with high objectness scores computed with the magnitude of the feature activations from the backbone are then pseudo-labeled as unknown objects with bounding boxes given by their corresponding regression branch predictions. Third, a novelty classification branch introduces the novelty class label so unknown instances can be distinguished from backgrounds. Finally, a foreground objectness branch is introduced to better separate foreground objects (known and unknown) from the background, which allows the knowledge to transfer from known to unknown objects. This transformer-based OWOD method achieves better results compared with ORE \cite{ORE}.

CAT \cite{CAT} is another pseudo-labeling-based OWOD method developed from OW-DETR \cite{OW-DETR}. The authors argue that it is human intrinsic to extract object localization and identification process. Thus, a shared cascaded transformer decoder is proposed to decouple object detection into two parts. Apart from the decoupled decoding structure, CAT introduces attention-driven pseudo-labeling combined with selective search to generate robust pseudo-labels for unknown objects self-adaptively. The self-adaptive pseudo-labeling scheme significantly improves the ability of CAT to retrieve unknown objects. Fast OWDETR \cite{chen2022fast} also develops from OW-DETR \cite{OW-DETR}. The author argues that the bounding box location of OW-DETR is imprecise, and the attention-driven pseudo-labeling employed by OW-DETR has high computational complexity. Thus, a bounding box refinement technique and logits-based simple pseudo-labeling scheme are deployed. 

Open World DETR proposed by N. Dong \textit{et al.} \cite{OpenWorldDETR} is also a pseudo-labeling-based OWOD method. After a pre-training of the model, the parameters of the feature extractor and regression head will be fixed to avoid known class biases. Then, a multi-view self-labeling scheme is adopted to generate pseudo-ground truth for unknown instances, and a swapped prediction mechanism is performed for image pairs with or without data augmentation to make consistent predictions for different views of the same image. Besides, selective search is also implemented to help propose other potential unknown regions. Finally, both exemplars replay and knowledge distillation strategies are used to alleviate the catastrophic forgetting problem of incremental learning.

\begin{figure*}
\centering
\includegraphics[scale=0.5]{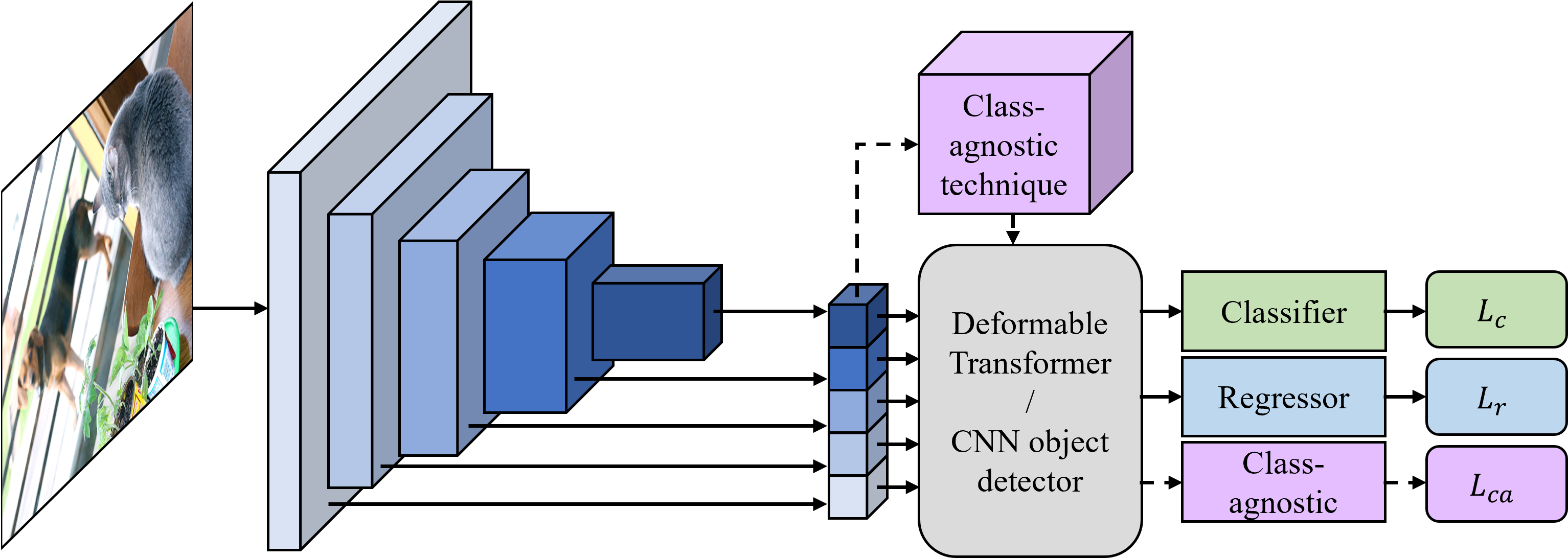}\caption{Framework of class-agnostic OWOD methods. The class-agnostic technique can be implemented after the feature extraction or head prediction process, as shown by the dashed lines.} \label{CA}
\vspace{-4mm}
\end{figure*}

\subsection{Class-agnostic methods}
Class-agnostic methods consider known and unknown objects as the same foreground objects. By separating the detection of objects and the identification of each instance, these methods use a class-agnostic object proposer to measure the objectness of proposed regions. As the class-agnostic object proposer is trained to learn the objectness rather than the classifier, no bias from known categories is introduced. A common framework of class-agnostic OWOD methods is shown in Fig. \ref{CA}.

Wu \textit{et al.} proposed a class-agnostic method called Two-branch Objectness-centric Open World Detection (2B-OCD) \cite{2B-OCD}, which adopts a class-agnostic objectness-centric calibrator to capture the objectness of both known and unknown instances. The authors argue that the classifier in previous works impedes generalization as it learns to classify whether a region belongs to predefined classes. 2B-OCD comprises an objectness-centric calibrator and a bias-guided detector with the same Faster R-CNN feature extractor. During the training process, the gradient of the objectness-centric calibrator will not be returned in order to reduce biases from known classes. In the reference stage, an objectness-centric affirmation is deployed to confirm the proposals with higher objectness confidence than the threshold and do not belong to known categories as unknown instances.

Probabilistic Objectness transformer-based open-world detector (PROB) \cite{PROB} is another method based on the class-agnostic strategy. Using Deformable DETR \cite{DDETR} as the feature extractor, PROB extends the baseline by adding an ``Unknown Object" class label and separating object and object class predictions. This allows the model to learn about objectness and object class probability separately. The objectness head estimates the probability of a query being an object, while the classification head classifies the query into known or unknown objects. A multivariate class-agnostic Gaussian distribution is used to parameterize the objectness probability in the query embedding space. In order to achieve better incremental learning results, PROB adopts an exemplar replay strategy to alleviate catastrophic forgetting. Objectness acquired in the previous module is used to select the instances of exemplar. The authors argue that instances with low objectness are expected to improve model performance on new objects, while instances with high objectness are expected to impede catastrophic forgetting. 

Like 2B-OCD \cite{2B-OCD}, OW-RCNN \cite{OW-RCNN} is another class-agnostic method using Faster R-CNN as a backbone. The authors propose three challenges of open-world object detection: class-agnostic region proposals, unknown-aware classification, and open-set error correction. Firstly, the Region Proposal Network (RPN) is trained to produce unknown-aware region proposals by predicting the distances from the anchor center to the edges of the ground truth bounding box. A regression-based localization quality head is trained to predict the centerness of the box head’s output. Secondly, unlike most other works, OW-RCNN classifies the unknown and background in the same category. By comparing the objectness score and per-class scores, OW-RCNN can determine whether the region belongs to a known class, an unknown class, or a background. Finally, Gaussian mixture models are utilized to determine the likelihood of the detection network’s classification output in order to reduce open-set errors. The models are used during the inference process to detect when the detection network has made an overconfident prediction.

A classification-free Object Localization Network (OLN) was proposed by D. Kim \textit{et al.} as another class-agnostic OWOD method. However, unlike the previous OWOD methods, this model only focuses on unknown detection rather than classifying all known and unknown categories. Thus, the unknown classification process is not considered in this paper. By replacing the commonly used classifier in object proposal methods with localization quality indicators such as Intersection of Union (IoU) and centerness, this method enables better generalization to unseen classes. The authors argue that the learned purely localization-based (location and shape) objectness cues can improve the generalization of object proposals, and a classification head is severely harmful. During the inference stage, the objectness score will be calculated as a geometric mean based on centernesss and IoU acquired above.

Wang \textit{et al.} \cite{RandBox} recently proposed another class-agnostic OWOD method using a random proposals generator (RandBox). Dynamic-$k$ matcher match between ground truth and proposals for known foreground objects. A self-defined matching score is used to select top $N$ unknown objects. Arguing that objectness score for unknown categories produced by RPN based on Faster R-CNN is penalized as it is trained on known categories. Mean activation of the ROI feature used by DETR-based methods suffers from unreliability and confusion between background and unknown categories. The author also proposed a confounding effect based on the casual model. The proposals generated are biased towards labeled data of known objects, which can lead to challenges like low recall of unknown categories. In their method, the region is generated randomly without influence from the training data, thereby eliminating the confounding effect.

\begin{figure*}
\centering
\includegraphics[scale=0.4]{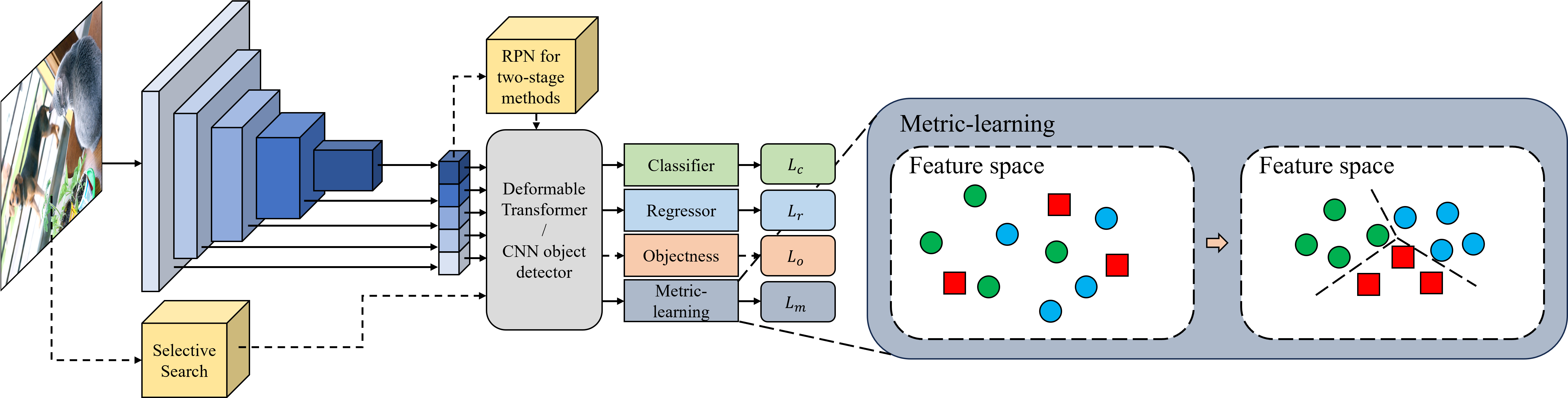}\caption{Framework of Metric-learning OWOD methods. Compared to pseudo-labeling-based methods, a metric-learning module is applied to help the model distinguish between known classes and unknown classes.} \label{ML}
\vspace{-4mm}
\end{figure*}

\subsection{Metric-learning methods}
Metric-learning OWOD methods generally treat the classification of unknown instances as a metic-learning process. By projecting the features of instances on an embedding feature space, a bunch of metric-learning techniques can be utilized to classify between known classes, unknown classes, and backgrounds. Most metric-learning methods use a common strategy to extract potential unknown instances and focus on distinguishing between known, unknown, and backgrounds. Some methods even extend to separate different unknown classes without ground truth labels, which is closer to real open-world settings. A summary structure of metric-learning OWOD methods is illustrated as Fig. \ref{ML}

\ym{As discussed in subsection \ref{sec3b}, ORE \cite{ORE} provides a novel OWOD solution based on contrastive clustering and energy-based unknown identification. Contrastive clustering is used to enforce class separation in the latent space. A prototype vector for each known class trained by a contrastive loss gradually evolves. In order to differentiate between known and unknown instances, the author proposed an energy-based classification head with Helmholtz free energy. However, ORE relies on weak supervision for unknown instances to estimate the distribution of unknown classes using a validation set.}

Revisiting Open World Object Detection \cite{RE-OWOD} (RE-OWOD) proposed by Zhao, \textit{et al.} utilizes a class-specific expelling classifier (CEC) to determine whether a proposal belongs to any known categories. Using the non-parametric selective search as an auxiliary proposal advisor (PAD) to confirm the proposals generated by the Region Proposal Network (RPN), the original RPN can provide more accurate potential unknown proposals for unknown classification later. The authors argue that DNN-based discriminative classifiers suffer from producing overconfidence, especially for OWOD settings, as there is no information for unknown classes. Thus, CEC is proposed to remove confusing instances from the predicted known class and reassign their class predictions. The class activation boundary of each known class will be calibrated by annotation information. If all classes expel the proposal, which means the proposal is not included in any of the known class activation areas, it will be predicted as the ``unknown" class.

Yu \textit{et al.} \cite{OCPL} proposed a class prototype-based metric learning method called OCPL: Open-world object detection via discriminative Class Prototype Learning. OCPL simply uses Faster R-CNN as the feature extractor, and RPN is used to generate potential regions. Proposal Embedding Aggregator is employed to optimize the prototype for each category using distance-based cross-entropy loss. In order to compress the range of known classes in feature space, the embedding space compressor is proposed to reduce the overlap between known and unknown distribution. Due to the high intra-class distance in the unknown category, a cosine similarity-based classifier forms a tighter clustering of instances in the same class. A threshold is used to ignore some detected instances with low classification scores.

Compared to the previous metric-learning OWOD methods, Unknown-Classified Open World Object Detection \cite{UC-OWOD} (UC-OWOD) can achieve different unknown class detection, which is closer to real OWOD settings. To achieve this, the unknown label-aware proposal is deployed to generate potential unknown regions, which is similar to ORE \cite{ORE}. UC-OWOD modifies the original single unknown classification head to unknown-discriminative classification heads, which can be used to distinguish between different unknown classes. Besides, similarity-based unknown classification is also deployed to determine whether unknown instance pairs are similar. Instance pairs are introduced gradually based on their difficulties in distinguishing to better cluster the instances. Finally, unknown clustering refinement is applied to improve the robustness of UC-OWOD using the soft assignment method followed by \cite{SoftAssign}. UC-OWOD also introduced UC-mAP and UC-Recall, which can better illustrate the features of unknown-discriminative OWOD methods.

\subsection{Other methods}
Apart from what has been included, there are also other OWOD methods that cannot be classified into any of the categories above. Some methods are even from related areas such as multi-modal object detection, out-of-distribution detection, etc. These methods show good potential for OWOD and, thus, are included in this review.

\ym{Ma \textit{et al.} \cite{ma2023annealing} introduce an Annealing-based Label-Transfer Learning framework for Open World Object Detection (OWOD) that utilizes object-level feature entanglement and a label-transfer approach to detect unknown objects without manual selection. The method incorporates a Sawtooth Annealing Scheduling to adjust decision boundaries dynamically between known and unknown classes, significantly improving detection accuracy for both. The study also proposes the Equilibrium Index, a metric designed to evaluate OWOD models by assessing their performance on known and unknown detections.}

Maaz \textit{et al.} \cite{MAVL} (MAVL) proposed a variant of ORE \cite{ORE} using a multimodal architecture called Multi-scale Attention ViT with Late fusion. MAVL uses a multi-scale deformable attention module to process multi-scale image features. RoBERTa \cite{liu2019roberta} model is used to extract features of corresponding text labels. Late fusion is implemented for vision-language fusion. In order to achieve Open World Object Detection, class-agnostic proposals obtained by MAVL are used in the ORE model as pseudo-labels. Although the comparison between single-modality and multi-modality is not fair because additional text information is introduced, MAVL provides another direction of OWOD methods.

Spatial-Temporal Unknown Distillation \cite{STUD} (STUD) is an unknown-aware object detection method from an out-of-distribution (OOD) detection area. The authors argue that OOD data can be selected effectively using energy score \cite{liu2020energy}. Therefore, using the labeled key frame and unlabeled reference frames, unknown object proposals can be identified and distilled accordingly. $L_{2}$ distance is adopted to measure the dissimilarity between unknown proposals and ground truth labels for spatial distillation. For temporal unknown distillation, feature vectors of unknown objects are concatenated. This method needs videos as the training data. Thus, standard evaluation protocols for OWOD methods are not available on STUD.

\ym{Liang \textit{et al.} introduce the ``Unknown Sniffer" (UnSniffer) \cite{liang2023unknown}, a novel framework for OWOD that enhances the detection of both known and unknown objects. Unlike traditional OWOD methods, UnSniffer leverages Generalized Object Confidence (GOC) scoring to distinguish objects from non-objects based on knowledge of known classes, thereby enhancing generalization to unknown objects. Furthermore, the framework replaces standard non-maximum suppression with a graph-based box determination method to optimize bounding box selection during inference. This method efficiently handles overlapping objects by clustering high-confidence proposals and selecting the most representative proposals from each cluster. To solve the common unknown object suppression problem in training, UnSniffer uses a negative energy suppression loss to distinguish non-object backgrounds, thereby reducing false positives. Furthermore, recognizing the shortcomings of existing benchmarks in evaluating unknown object detection, the authors propose the Unknown Object Detection Benchmark (UOD-Benchmark), which includes a finely annotated dataset designed to test unknown object detection performance. The details of such benchmark is also introduced in section \ref{sec4a}}

Detecting the open-world objects with the help of the ``Brain" proposed by S. Ma \textit{et al.} \cite{DOWB} (DOWB) uses an extra large-scale pre-trained grounded language-image model as the ``Brain" of the OWOD model. Apart from the classical OWOD model with unknown proposal and unknown classification modules, DOWB adopts an assistant module with the Grounded Language-Image Pre-training \cite{GLIP} (GLIP) model to provide unknown instance proposals. The authors argue that leveraging the GLIP model is non-trivial because the unknown labels impair the model’s learning of known objects. To alleviate these problems, they propose a down-weight loss function and decoupled detection structure. The down-weight training strategy leverages the generated identification confidence to generate soft labels and down-weight the unknown training loss. The training loss function comprises several components, including regression loss, box score loss, classification loss, etc. However, the pre-trained model introduces extra knowledge from the large-scale dataset for pertaining. 
\section{Datasets and Evaluation Metrics}\label{sec4}
\subsection{Datasets}\label{sec4a}
\subsubsection{MS-COCO}
MS-COCO dataset \cite{Lin2014MicrosoftCC} is one of the most commonly used datasets for object detection tasks. This dataset comprises 164k images, among which 83k are for training, 40k for validation, and 41k for testing. These samples are labeled in 80 different categories. Due to its large size and diverse range of object categories, MS-COCO has become a useful resource for developing and testing object detection algorithms, including OWOD, in a variety of real-world scenarios.


\subsubsection{Pascal VOC} PASCAL VOC \cite{Everingham2010ThePV} is another popular benchmark  for object detection tasks. It contains images of objects in 20 different categories (a subset of the MS-COCO label set), such as person, car, cat, and dog, etc. Each image is annotated with object bounding boxes, object class labels, and object segmentation masks. The dataset includes indoor and outdoor scenes, and the objects are presented in various poses, scales, and orientations.

\ld{\subsubsection{The new datasets} To further label comprehensive unknown objects from images, COCO-OOD and COCO-Mixed \cite{liang2023unknown} are proposed based on MS-COCO \cite{Lin2014MicrosoftCC}. First, COCO-OOD arguments the original MS-COCO categories into 1655 unknown objects. There are only 504 images with unknown objects in this dataset. Meanwhile, COCO-Mixed contains 2658 known objects (including original COCO annotations) and 2533 unknown objects. There are 897 images in total. The fine-grained annotations make COCO-Mixed a more challenging dataset.}
\subsubsection{Dataset splits in OWOD task}
In the task of open-world object detection, the dataset is divided into several splits based on two strategies.
First, in the original OWOD task, Joseph \textit{et al.} \cite{ORE} integrates the MS-COCO dataset with the PASCAL VOC dataset to provide more samples, called OWOD split. Specifically, all the classes and the corresponding samples are grouped into a set of non-overlapping tasks $\{T_1, \cdots, T_t\}$. Classes from the PASCAL VOC dataset are treated as task $T_1$. The other classes are grouped into tasks by semantic drifts. The detailed information of the dataset is listed in \ref{tab:data_split}.

In the latest OWOD task, Gupta \textit{et al.} \cite{OW-DETR} proposed a new strategy by splitting the categories across super-classes, called MS-COCO split. Specifically, object classes are grouped into the same tasks by semantic meanings. for example, \textit{trucks} and \textit{vehicles} that belong to different tasks in the combined dataset are grouped into the same super-class task: \textit{Animals, Person, Vehicles}. More detailed statistics are given in Table \ref{tab:data_split2}.


\begin{table}
\centering
\caption{Task composition in OWOD split on MS-COCO and Pascal VOC datasets in \cite{ORE}. The semantics of each task and the number of images and instances (objects) of each task are represented.}
\vspace{-2mm}
\resizebox{0.48\textwidth}{!}{%
\begin{tabular}{@{}l|cccc@{}}
\hline
\toprule
 & Task 1 & Task 2 & Task 3 & Task 4 \\ \midrule
Semantic split & \begin{tabular}[c]{@{}c@{}}VOC \\ Classes\end{tabular} & \begin{tabular}[c]{@{}c@{}}Outdoor, Accessories, \\ Appliance, Truck\end{tabular} & \begin{tabular}[c]{@{}c@{}}Sports, \\ Food\end{tabular} & \begin{tabular}[c]{@{}c@{}}Electronic, Indoor, \\ Kitchen, Furniture\end{tabular} \\\midrule
\hline
\# training images & 16,551 & 45,520 & 39,402 & 40,260 \\

\# test images & 4,952 & 1,914 & 1,642 & 1,738 \\

\# train instances & 47,223 & 113,741 & 114,452 & 138,996 \\

\# test instances & 14,976 & 4,966 & 4,826 & 6,039 \\ \bottomrule
\hline
\end{tabular}%
}
\label{tab:data_split}
\vspace{-2mm}
\end{table}

\begin{table}
\centering
\caption{Task composition in MS-COCO split on MS-COCO dataset in \cite{OW-DETR}. The semantics of each task and the number of images and instances (objects) of each task are represented.}
\vspace{-2mm}
\resizebox{0.48\textwidth}{!}{%
\begin{tabular}{@{}l|cccc@{}}
\hline
\toprule
 & Task 1 & Task 2 & Task 3 & Task 4 \\ \midrule
Semantic split & \begin{tabular}[c]{@{}c@{}}Animals,\\ Person,
Vehicles\end{tabular} & \begin{tabular}[c]{@{}c@{}}
Appliances, Accessories,\\
Outdoor, Furniture\end{tabular} & \begin{tabular}[c]{@{}c@{}}Sports, \\ Food\end{tabular} & \begin{tabular}[c]{@{}c@{}}Electronic, \\Indoor,  Kitchen\end{tabular} \\\midrule
\hline
\# training images &~89,490~  & ~55,870~  & ~39,402~  & ~38,903~  \\

\# test images & ~3,793~   & ~2,351~   & ~1,642~   & ~1,691~   \\

\# train instances & ~421,243~ & ~163,512~ & ~114,452~ & ~160,794~\\

\# test instances & ~17,786~  & ~7,159~   & ~4,826~   & ~7,010 \\ \bottomrule
\hline
\end{tabular}%
}
\label{tab:data_split2}
\vspace{-4mm}
\end{table}

\subsection{Evaluation Metrics}\label{sec4b}
Here, we first introduce the widely used evaluation metrics in the OWOD task, including Wilderness Impact (WI), Absolute Open-Set Error (A-OSE), Mean Average Precision (mAP), and Unknown-recall (U-recall). The first two metrics are designed to evaluate the effectiveness of object detection models in handling unknown objects. In addition, some methods also report other metrics to evaluate the open world ability of their models, such as Unknown Mean Average Precision (UC-mAP) in \cite{RE-OWOD}, Unknown Detection Recall (UDR) and Unknown Detection Prediction (UDP) in \cite{RE-OWOD}, and Unknown F1-Score defined in \cite{liang2023unknown}.

\textbf{Wilderness Impact (WI)} \cite{Dhamija2020TheOE} evaluates the impact of unknown objects on the detector’s performance. The WI metric can be calculated based on the precisions of known and unknown objects detected by the algorithm as:
\begin{equation}
   WI = \frac{P_{K}}{P_{K \cup U}} - 1,
\end{equation}
where $P_{K}$ refers to the precision of the detector evaluated on known objects, and $P_{K \cup U}$ denotes the precision of the detector evaluated on all objects, including both known and unknown objects. A lower value in the WI metric means a better result, indicating that the algorithm's precision does not drop significantly when unknown objects are added to the test set. A low WI score implies that the algorithm is robust and can generalize well to new and unseen objects, which is an important aspect of object detection in real-world applications.

\textbf{Absolute Open-Set Error (A-OSE)} \cite{Miller2017DropoutSF} is also used to evaluate the performance of a detector in identifying unknown objects. This metric reports the total number of unknown objects that are classified as any of the known objects, providing a measure of the algorithm's ability to distinguish between known and unknown objects. A low A-OSE score indicates that the algorithm can effectively distinguish between known and unknown objects, while a high A-OSE score indicates that the algorithm is more likely to misclassify unknown objects as known objects.

\textbf{Mean Average Precision (mAP)} is a commonly used evaluation metric for object detection and recognition models. This metric can be calculated as:
\begin{equation}
    mAP=\frac{1}{N} \sum_{i=1}^{N} AP_i,
    \label{map}
\end{equation}
where $N$ denotes the number of object classes. $AP_i$ represents the average precision of class $i^{th}$ under a certain Intersection over Union (IoU) threshold value (usually IoU threshold of 0.5). In specific, the threshold is varied from 0 to 1. At each threshold value, the precision and recall values are first calculated based on the model's predictions and ground truth labels. Then, the area under the precision-recall curve is calculated to obtain the AP value for Class $i$. Overall, mAP is a useful metric for evaluating the performance of object detection models, providing insight into the model's ability to accurately and consistently detect objects.

\textbf{Unknown Mean Average Precision (UC-mAP)}\cite{UC-OWOD} is a variant of mAP, with automatic unknown category matching as:

\begin{equation}
    UC \text{-} mAP(y, \hat{y})= \max_{p \in P} mAP( p(\hat{y}), y),
\end{equation}
where $y$ is the ground truth and $\hat{y}$ is the output of the prediction. P denotes all permutations in 1 to M, where M is the number of unknown categories. mAP can be calculated by Equation \ref{map}.

\textbf{Unknown-Recall (U-Recall)} measures the ability of a classifier to correctly identify unknown or novel classes. It is calculated by 
the proportion of unknown samples that are correctly detected as unknown by the model as:
\begin{equation}
    U \text{-} Recall=\frac{TP_u}{TP_u + FN_u},
\end{equation}
where $TP_u$ denotes the number of correctly identified unknown samples (true positive), and $FN_U$ denotes the number of unknown samples that are not detected (False negative).

\textbf{Unknown Detection Recall (UDR)} \cite{RE-OWOD} evaluates the accurate localization of unknown categories. It can be calculated as:
\begin{equation}
    UDR=\frac{TP_u+FN_{u}^{*}}{TP_u+FN_u},
\end{equation}
where $FN_{u}^{*}$ denotes the number of ground-truth boxes recalled by the predicted bounding boxes that are misclassified.
\textbf{Unknown Detection Precision (UDP) }\cite{RE-OWOD} measures the accurate classification of
all localized unknown instances. This can be calculated by:
\begin{equation}
    UDP=\frac{TP_u}{TP_u + FN_{u}^{*}}.
\end{equation}
Compared to U-Recall, UDR and UDP consider both localization and classification applications. Also, $FN_{u}^{*}$ are considered as correct recall.

\textbf{Unknown F1-Score (U-F1)} \cite{liang2023unknown} reports the harmonic mean of U-Recall and Precision Rate of Unknown (U-Pre) for a comprehensive comparison, which can be calculated as:
\begin{equation}
    U\text{-}F1=\frac{2\times U\text{-}Pre\times U\text{-}Recall}{U\text{-}Pre+ U\text{-}Recall},
\end{equation}
where $U\text{-}Pre$ is calculated as:
\begin{equation}
    U\text{-}Pre=\frac{TP_{u}}{TP_{u}+FP_{u}} 
\end{equation}

\subsection{Comparative Results}
In this subsection, we present the comparative results of different OWOD methods in previous sections and summarize their performance under two evaluation protocols. Four commonly used evaluation metrics are selected to demonstrate different features of methods.

\subsubsection{Comparison under OWOD split}
In Table \ref{tab:OWOD}, we evaluate most state-of-the-art methods on OWOD split composed of PASCAL VOC and MS-COCO datasets as described in section \ref{sec4b}. The results of ORE\cite{ORE} are from OW-DETR \cite{OW-DETR}, which excluded the held-out validation set EBUI. Note that the comparison is not completely fair because of the introduction of extra training information. Many OWOD methods with extra information \cite{MAVL,DOWB} perform better on unknown recall metrics. Specifically, the DOWB \cite{DOWB} method implements a large-scale image-text pre-trained model that introduces additional information from the pre-trained dataset. The MAVL \cite{MAVL} method also uses extra text modality information to enhance the performance of unknown detection. The results of some methods in the previous section are not applicable due to the different evaluation protocols and training methods. 

Models are categorized into Faster R-CNN and Deformable DETR (D-DETR) based on the backbone network they use for a general comparison. The mAP of the current known, termed ``CK", stands for the mean average precision of learned categories in the current task. Therefore, the mAP score of previously known, termed ``PK", is the mean average precision of learned categories in previous tasks. ``Both" is the weighted average of all the known categories. 

\begin{table*}[htbp]
  \centering
  \caption{Comparison on OWOD split}
  \vspace{-2mm}
  \resizebox{\textwidth}{!}{
    \begin{tabular}{l|c|cc|cccc|cccc|ccc}
    \hline
    \multicolumn{2}{c|}{Task IDs} & \multicolumn{2}{c|}{Task 1} & \multicolumn{4}{c|}{Task 2}   & \multicolumn{4}{c|}{Task 3}   & \multicolumn{3}{c}{Task 4} \\
    \hline
    \multicolumn{1}{c|}{\multirow{2}[2]{*}{Metrics}} & \multirow{2}[2]{*}{Backbone} & \multicolumn{1}{c}{\cellcolor[rgb]{ 1,  1,  .8}U-Recall} & mAP   & \cellcolor[rgb]{ 1,  1,  .8}U-Recall & \multicolumn{3}{c|}{mAP} & \cellcolor[rgb]{ 1,  1,  .8}U-Recall & \multicolumn{3}{c|}{mAP} & \multicolumn{3}{c}{mAP} \\
          &     & \cellcolor[rgb]{ 1,  1,  .8} & \multicolumn{1}{c|}{CK} & \cellcolor[rgb]{ 1,  1,  .8} & \multicolumn{1}{c}{PK} & \multicolumn{1}{c}{CK} & \multicolumn{1}{c|}{Both} & \cellcolor[rgb]{ 1,  1,  .8} & \multicolumn{1}{c}{PK} & \multicolumn{1}{c}{CK} & \multicolumn{1}{c|}{Both} & \multicolumn{1}{c}{PK} & \multicolumn{1}{c}{CK} & \multicolumn{1}{c}{Both} \\
    \hline
    ORE \cite{ORE}  & \multirow{7}[2]{*}{\begin{tabular}[c]{@{}c@{}}Faster\\ R-CNN\end{tabular}}  & \cellcolor[rgb]{ 1,  1,  .8}4.9 & 56.0  & \cellcolor[rgb]{ 1,  1,  .8}2.9 & 52.7  & 26.0  & 39.4  & \cellcolor[rgb]{ 1,  1,  .8}3.9 & 38.2  & 12.7  & 29.7  & 29.6  & 12.4  & 25.3 \\
    UC-OWOD \cite{UC-OWOD} &    & \cellcolor[rgb]{ 1,  1,  .8}2.4 & 50.7  & \cellcolor[rgb]{ 1,  1,  .8}3.4 & 33.1  & 30.5  & 31.8  & \cellcolor[rgb]{ 1,  1,  .8}8.7 & 28.8  & 16.3  & 24.6  & 25.6  & 12.9  & 23.2 \\
    OCPL \cite{OCPL} &    & \cellcolor[rgb]{ 1,  1,  .8}8.3 & 56.6  & \cellcolor[rgb]{ 1,  1,  .8}7.7 & 50.7  & 27.5  & 39.1  & \cellcolor[rgb]{ 1,  1,  .8}11.9 & 38.6  & 14.7  & 30.7  & 30.8  & 14.4  & 26.7 \\
    2B-OCD \cite{2B-OCD} &    & \cellcolor[rgb]{ 1,  1,  .8}12.1 & 56.4  & \cellcolor[rgb]{ 1,  1,  .8}9.4 & 51.6  & 25.3  & 38.5  & \cellcolor[rgb]{ 1,  1,  .8}11.7 & 37.2  & 13.2  & 29.2  & 30.0  & 13.3  & 25.8 \\
    RE-OWOD \cite{RE-OWOD} &    & \cellcolor[rgb]{ 1,  1,  .8}9.1 & 59.7  & \cellcolor[rgb]{ 1,  1,  .8}9.9 & 54.1  & 37.3  & 45.6  & \cellcolor[rgb]{ 1,  1,  .8}11.4 & 43.1  & 24.6  & 37.6  & 38.0  & 28.7  & 35.7 \\
    OW-RCNN \cite{OW-RCNN} &    & \cellcolor[rgb]{ 1,  1,  .8}37.7 & 63.0  & \cellcolor[rgb]{ 1,  1,  .8}39.9 & 48.8  & 41.7  & 45.2  & \cellcolor[rgb]{ 1,  1,  .8}43.0 & 45.2  & 31.7  & 40.7  & 40.3  & 28.8  & 37.4 \\
    RandBox \cite{RandBox} &    & \cellcolor[rgb]{ 1,  1,  .8}10.6 & 61.8  & \cellcolor[rgb]{ 1,  1,  .8}6.3 & -  & -  & 45.3  & \cellcolor[rgb]{ 1,  1,  .8}7.8 & -  & -  & 39.4  & -  & -  & 35.4 \\
    Ma et al. (Faster R-CNN) \cite{ma2023annealing} &    & \cellcolor[rgb]{ 1,  1,  .8}12.8 & 56.7  & \cellcolor[rgb]{ 1,  1,  .8}5.0 & -  & -  & 40.6  & \cellcolor[rgb]{ 1,  1,  .8}9.8 & -  & -  & 32.0  & -  & 27.0  & 37.4 \\
    \hline
    OW-DETR \cite{OW-DETR} & \multirow{7}[2]{*}{D-DETR} & \cellcolor[rgb]{ 1,  1,  .8}7.5 & 59.2  & \cellcolor[rgb]{ 1,  1,  .8}6.2 & 53.6  & 33.5  & 42.9  & \cellcolor[rgb]{ 1,  1,  .8}5.7 & 38.3  & 15.8  & 30.8  & 31.4  & 17.1  & 27.8 \\
    Fast-OWDETR \cite{chen2022fast} &    & \cellcolor[rgb]{ 1,  1,  .8}9.2 & 56.6  & \cellcolor[rgb]{ 1,  1,  .8}8.8 & 51.3  & 28.6  & 39.4  & \cellcolor[rgb]{ 1,  1,  .8}7.8 & 39.2  & 15.7  & 32.2  & 28.2  & 11.4  & 25.0 \\
    Open World DETR \cite{OpenWorldDETR} &    & \cellcolor[rgb]{ 1,  1,  .8}21.0 & 59.9  & \cellcolor[rgb]{ 1,  1,  .8}15.7 & 51.8  & 36.4  & 44.1  & \cellcolor[rgb]{ 1,  1,  .8}17.4 & 38.9  & 24.7  & 34.2  & 32.0  & 19.7  & 29.0 \\
    MAVL \cite{MAVL}  &    & \cellcolor[rgb]{ 1,  1,  .8}50.1 & 64.0  & \cellcolor[rgb]{ 1,  1,  .8}49.5 & 61.6  & 30.8  & 46.2  & \cellcolor[rgb]{ 1,  1,  .8}50.9 & 43.8  & 22.7  & 36.8  & 36.2  & 20.6  & 32.3 \\
    PROB \cite{PROB}  &    & \cellcolor[rgb]{ 1,  1,  .8}19.4 & 59.5  & \cellcolor[rgb]{ 1,  1,  .8}17.4 & 55.7  & 32.2  & 44.0  & \cellcolor[rgb]{ 1,  1,  .8}19.6 & 43.0  & 22.2  & 36.0  & 35.7  & 18.9  & 31.5 \\
    CAT \cite{CAT} &    & \cellcolor[rgb]{ 1,  1,  .8}21.8 & 59.9  & \cellcolor[rgb]{ 1,  1,  .8}18.6 & 54.0  & 33.6  & 43.8  & \cellcolor[rgb]{ 1,  1,  .8}23.9 & 42.1  & 19.8  & 34.7  & 35.1  & 17.1  & 30.6 \\
    DOWB \cite{DOWB}  &    & \cellcolor[rgb]{ 1,  1,  .8}39.0 & 56.8  & \cellcolor[rgb]{ 1,  1,  .8}36.7 & 52.3  & 28.3  & 40.3  & \cellcolor[rgb]{ 1,  1,  .8}36.1 & 36.9  & 16.4  & 30.1  & 31.0  & 14.7  & 26.9 \\
    Ma et al. (DETR) \cite{ma2023annealing} &    & \cellcolor[rgb]{ 1,  1,  .8}12.8 & 56.7  & \cellcolor[rgb]{ 1,  1,  .8}5.0 & -  & -  & 40.6  & \cellcolor[rgb]{ 1,  1,  .8}9.8 & -  & -  & 32.0  & -  & 27.0  & 37.4 \\
    \hline
    \end{tabular}%
    }
  \label{tab:OWOD}%
  \vspace{-2mm}
\end{table*}%

\begin{table*}[htbp]
  \centering
  \caption{Comparison on MS-COCO split}
  \vspace{-2mm}
  \resizebox{\textwidth}{!}{
    \begin{tabular}{l|c|cc|cccc|cccc|ccc}
    \hline
    \multicolumn{2}{c|}{Task IDs} & \multicolumn{2}{c|}{Task 1} & \multicolumn{4}{c|}{Task 2}   & \multicolumn{4}{c|}{Task 3}   & \multicolumn{3}{c}{Task 4} \\
    \hline
    \multicolumn{1}{c|}{\multirow{2}[2]{*}{Metrics}} & \multirow{2}[2]{*}{Backbone} & \multicolumn{1}{c}{\cellcolor[rgb]{ 1,  1,  .8}U-Recall} & mAP & \cellcolor[rgb]{ 1,  1,  .8}U-Recall & \multicolumn{3}{c|}{mAP} & \cellcolor[rgb]{ 1,  1,  .8}U-Recall & \multicolumn{3}{c|}{mAP} & \multicolumn{3}{c}{mAP} \\
          &     & \cellcolor[rgb]{ 1,  1,  .8} & \multicolumn{1}{c|}{CK} & \cellcolor[rgb]{ 1,  1,  .8} & \multicolumn{1}{c}{PK} & \multicolumn{1}{c}{CK} & \multicolumn{1}{c|}{Both} & \cellcolor[rgb]{ 1,  1,  .8} & \multicolumn{1}{c}{PK} & \multicolumn{1}{c}{CK} & \multicolumn{1}{c|}{Both} & \multicolumn{1}{c}{PK} & \multicolumn{1}{c}{CK} & \multicolumn{1}{c}{Both} \\
    \hline
     ORE \cite{ORE}  & Faster  & \cellcolor[rgb]{ 1,  1,  .8}1.5 & 61.4  & \cellcolor[rgb]{ 1,  1,  .8}3.9 & 56.5  & 26.1  & 40.6  & \cellcolor[rgb]{ 1,  1,  .8}3.6 & 38.7  & 23.7  & 33.7  & 33.6  & 26.3  & 31.8 \\
     OW-RCNN \cite{OW-RCNN} &  R-CNN  & \cellcolor[rgb]{ 1,  1,  .8}23.9 & 68.9  & \cellcolor[rgb]{ 1,  1,  .8}33.3 & 49.6  & 36.7  & 41.9  & \cellcolor[rgb]{ 1,  1,  .8}40.8 & 42.3  & 30.8  & 38.5  & 39.4  & 32.2  & 37.7 \\
     \hline
     OW-DETR \cite{OW-DETR} & \multirow{4}[2]{*}{D-DETR} & \cellcolor[rgb]{ 1,  1,  .8}5.7 & 71.5  & \cellcolor[rgb]{ 1,  1,  .8}6.2 & 62.8  & 27.5  & 43.8  & \cellcolor[rgb]{ 1,  1,  .8}6.9 & 45.2  & 24.9  & 38.5  & 38.2  & 28.1  & 33.1 \\
     PROB \cite{PROB} &     & \cellcolor[rgb]{ 1,  1,  .8}19.4 & 59.5  & \cellcolor[rgb]{ 1,  1,  .8}17.4 & 55.7  & 32.2  & 44.0  & \cellcolor[rgb]{ 1,  1,  .8}19.6 & 43.0  & 22.2  & 36.0  & 35.7  & 18.9  & 31.5 \\
     CAT \cite{CAT}  &     & \cellcolor[rgb]{ 1,  1,  .8}24.0 & 74.2  & \cellcolor[rgb]{ 1,  1,  .8}23.0 & 67.6  & 35.5  & 50.7  & \cellcolor[rgb]{ 1,  1,  .8}24.6 & 51.2  & 32.6  & 45.0  & 45.4  & 35.1  & 42.8 \\
     DOWB \cite{DOWB} &     & \cellcolor[rgb]{ 1,  1,  .8}60.9 & 69.4  & \cellcolor[rgb]{ 1,  1,  .8}60.0 & 63.8  & 26.9  & 44.4  & \cellcolor[rgb]{ 1,  1,  .8}58.6 & 46.2  & 28.0  & 40.1  & 41.8  & 29.6  & 38.7 \\
    \hline
    \end{tabular}%
    }
  \label{tab:MS-COCO}%
\end{table*}%

\subsubsection{Comparison under MS-COCO split} 

We compare different methods on MS-COCO split as Table \ref{tab:MS-COCO} proposed in OW-DETR \cite{OW-DETR} described in section 4b. Only several models provide their results on the MS-COCO split. The results of ORE are from OW-DETR \cite{OW-DETR}, which excluded the held-out validation set EBUI, and some methods \cite{DOWB} introduce additional information.

Compared to the OWOD split, the MS-COCO split only uses the MS-COCO dataset as the training and testing set and introduces all the classes in a super-category in one task to mitigate data leakage across tasks and make it more challenging for OWOD. Thus, the results of the MS-COCO split are recommended to be reported by every OWOD method in the future. The settings of ``Current known", ``Previously known" and ``Both" are the same with OWOD splits described in the previous subsection.

\subsubsection{Comparison of other evaluation metrics}
The comparison of state-of-the-art OWOD methods under other evaluation metrics is shown in Table \ref{tab:EvaluationMetrics}. Apart from Unknown Recall ``U-Recall", Wilderness Impact ``WI" and Absolute-Open Set Error ``A-OSE" are reported by most OWOD methods. WI and A-OSE are defined in Section 4b. The up arrow ``$\uparrow$" indicates that the higher the value, the better the performance. The down arrow ``$\downarrow$" means a lower value has a better result.

According to the results of different OWOD methods, the WI and A-OSE evaluation metrics are not closely correlated with the U-Recall indicator. Low values of WI and A-OSE do not guarantee a good result of Unknown Recall. The inner relationship of these evaluation metrics needs to be investigated further. 

\begin{table*}[htbp]
  \centering
  \vspace{-0.2cm}
  \caption{Comparison under Open World Evaluation Metrics on MS-COCO split}
  \vspace{-2mm}
  \resizebox{\textwidth}{!}{
    \begin{tabular}{l|c|ccc|ccc|ccc}
    \hline
    \multicolumn{2}{c|}{Task IDs} & \multicolumn{3}{c|}{Task 1} & \multicolumn{3}{c|}{Task 2} & \multicolumn{3}{c}{Task 3} \\
    \hline
    \multicolumn{1}{c|}{\multirow{2}[2]{*}{Metrics}} & \multirow{2}[2]{*}{Backbone} & U-Recall ($\uparrow$) & WI ($\downarrow$)   & A-OSE ($\downarrow$) & U-Recall ($\uparrow$)& WI ($\downarrow$)   & A-OSE ($\downarrow$) & U-Recall ($\uparrow$)& WI ($\downarrow$)   & A-OSE ($\downarrow$)\\
          &       &       &       &       &       &       &       &       &  \\
    \hline
    ORE \cite{ORE}  & \multirow{7}[2]{*}{\begin{tabular}[c]{@{}c@{}}Faster\\ R-CNN\end{tabular}} & 4.9   & 0.0621 & 10459 & 2.9   & 0.0282 & 10445 & 3.9   & 0.0021 & 7990 \\
    UC-OWOD \cite{UC-OWOD} &    & 2.4   & 0.0136 & 9294  & 3.4   & 0.0116 & 5602  & 8.7   & 0.0073 & 3801 \\
    OCPL \cite{OCPL} &    & 8.3   & 0.0423 & 5670  & 7.7   & 0.0220 & 5690  & 11.9  & 0.0162 & 5166 \\
    2B-OCD \cite{2B-OCD} &    & 12.1  & 0.0481 & -     & 9.4   & 0.0160 & -     & 11.7  & 0.0137 & - \\
    RE-OWOD \cite{RE-OWOD} &    & 9.1   & 0.0449 & -     & 9.9   & 0.0331 & -     & 11.4  & 0.0241 & - \\
    OW-RCNN \cite{OW-RCNN} &    & 37.7  & 0.0524 & 6957  & 39.9  & 0.0233 & 2487  & 43.0  & 0.0165 & 1820 \\
    RandBOx \cite{RandBox} &    & 10.6  & 0.0240 & 4498 & 6.3  & 0.0078 & 1880  & 7.8  & 0.0054 & 1452 \\
    \hline
    OW-DETR \cite{OW-DETR} & \multirow{4}[2]{*}{D-DETR} & 7.5   & 0.0571 & 10240 & 6.2   & 0.0278 & 8441  & 5.7   & 0.0156 & 6803 \\
    CAT \cite{CAT}  &    & 21.8  & 0.0581 & 7070  & 18.6  & 0.0263 & 5902  & 23.9  & 0.0177 & 5189 \\
    Open World DETR \cite{OpenWorldDETR} &    & 21.0  & 0.0549 & 5909  & 15.7  & 0.0210 & 4378  & 17.4  & 0.0133 & 2895 \\
    PROB \cite{PROB} &    & 19.4  & 0.0569 & 5195  & 17.4  & 0.0344 & 6452  & 19.6  & 0.0151 & 2641 \\
    \hline
    \end{tabular}%
  }
  \label{tab:EvaluationMetrics}%
  \vspace{-4mm}
\end{table*}%

\begin{table}[htbp]
  \centering
  \vspace{-2mm}
  \caption{Comparison under Incremental Learning Object Detection Evaluation Protocol}
  \vspace{-2mm}
    \begin{tabular}{lccc}
    \hline
    10 + 10 setting & old classes & new classes & final mAP \\
    \hline
    ILOD \cite{ILOD} & 63.2  & 63.2  & 63.2 \\
    Faster ILOD \cite{FasterILOD} & 69.8  & 54.5  & 62.1 \\
    ORE - EBUI \cite{OW-DETR} & 60.4  & 68.8  & 64.5 \\
    OW-DETR \cite{OW-DETR} & 63.5  & 67.9  & 65.7 \\
    PROB \cite{PROB} & 66.0  & 67.2  & 66.5 \\
    Open World DETR \cite{OpenWorldDETR} & 67.8  & 70.9  & 69.3 \\
    CAT \cite{CAT}  & 67.9  & 67.4  & 67.7 \\
    DOWB \cite{DOWB} & 69.3  & 67.9  & 68.6 \\
    \hline
          &       &       &  \\
    \hline
    15 + 5 setting & old classes & new classes & final mAP \\
    \hline
    ILOD \cite{ILOD} & 68.3  & 58.4  & 65.8 \\
    Faster ILOD \cite{FasterILOD} & 71.6  & 56.9  & 67.9 \\
    ORE - EBUI \cite{OW-DETR} & 71.8  & 58.7  & 68.5 \\
    OW-DETR \cite{OW-DETR} & 72.2  & 59.8  & 69.4 \\
    PROB \cite{PROB} & 73.2  & 60.8  & 70.1 \\
    Open World DETR \cite{OpenWorldDETR} & 74.7  & 56.9  & 70.2 \\
    CAT \cite{CAT}  & 75.6  & 59.3  & 71.5 \\
    DOWB \cite{DOWB} & 76.6  & 58.8  & 72.1 \\
    \hline
          &       &       &  \\
    \hline
    19 + 1 setting & old classes & new classes & final mAP \\
    \hline
    ILOD \cite{ILOD} & 68.5  & 62.7  & 68.2 \\
    Faster ILOD \cite{FasterILOD} & 68.9  & 61.1  & 68.5 \\
    ORE - EBUI \cite{OW-DETR} & 69.4  & 60.1  & 68.8 \\
    OW-DETR \cite{OW-DETR} & 70.2  & 62.0  & 70.2 \\
    PROB \cite{PROB} & 73.9  & 48.5  & 72.6 \\
    Open World DETR \cite{OpenWorldDETR} & 76.6  & 64.4  & 76.0 \\
    CAT \cite{CAT}  & 74.4  & 61.1  & 73.8 \\
    DOWB \cite{DOWB} & 73.4  & 63.0  & 72.9 \\
    \hline
    \end{tabular}%
  \label{tab:ILOD}%
  \vspace{-4mm}
\end{table}%

\subsubsection{Comparison of incremental learning performance}
In ORE \cite{ORE}, Joseph \textit{et al.} proposed that ORE reduces the confusion of an unknown object being classified as a known object, and performs favorably well on incremental object detection. Thus, a result of incremental object detection under standard protocol was reported in ORE, and some following methods also present the results accordingly. The results of incremental learning are summarized in Table. \ref{tab:ILOD}.

Following the standard protocol \cite{ILOD,FasterILOD} used in incremental object detection, we evaluate the incremental learning performance of OWOD methods on different splits of Pascal VOC 2007. The models are trained on some (10, 15 or 19) classes and incrementally learn the other (10, 5 or 1) classes. The ``old classes" stands for the mAP of classes that are used as training set, and ``new classes" denoted the mAP of incremental learning classes. The ``final mAP" is the mean average of all 20 classes.
\section{Challenges and Future Trends}\label{sec5}
In this section, we discuss the challenges of Open World Object Detection and point out some potential directions and trends of OWOD research in the future. 

\subsection{Challenges of OWOD}
Based on the OWOD framework and current research status, we address the challenges and problems of OWOD as follows. 

\subsubsection{Unknown proposal}
The unknown instance proposal is the first key challenge in OWOD. As the training data only contains the labeled bounding boxes of known objects, the unlabeled region may contain obscured known objects, unknown objects, or background (no object). Thus, the proposal of unknown instances can be affected by labeled known training data, which can also be called ``bias from known''. 

To solve this issue, pseudo-labeling is used to label the potential unknown instances based on different kinds of objectness scores by many current works \cite{ORE, OW-DETR, chen2022fast, CAT, RE-OWOD, OCPL, UC-OWOD}. However, the training of objectness scores is based on the class-agnostic instance proposal trained on known objects. The ``bias from known'' issue is not solved completely. 

Apart from pseudo-labeling, class-agnostic OWOD methods minimizes the impact of known categories for object proposal modules by considering the known and unknown objects as the same foreground objects and separating the object proposal and detection branches. Several techniques are proposed such as no-gradient-return in 2B-OCD \cite{2B-OCD}, probabilistic separation in PROB \cite{PROB}, and localization-quality-based centerness head in OW-RCNN \cite{OW-RCNN}. Although training data of known classes are still inevitably involved, the results have generally improved compared to pseudo-labeling methods.

\subsubsection{Unknown classification}
Unlike classical object detection task which tries to classify between known classes and backgrounds, OWOD task deals with the additional unknown class. The introduction of the unknown class will generate two additional requirements for the OWOD methods. First, the classifier needs to classify the instances as known, background, or unknown. Secondly, according to \cite{Miller2017DropoutSF}, the introduction of unknown instances will lead to open-set errors, i.e. unknown objects are detected and misclassified as known objects.

In order to achieve unknown classification, most OWOD methods \cite{ORE, OW-DETR} just simply add another unknown classification head. Some methods add other techniques such as decoupled decoder in CAT \cite{CAT} and Gaussian mixture model in OW-RCNN \cite{OW-RCNN} apart from additional classification head to reduce open-set errors.

Based on methods in the metric learning area, some metric-learning-based unknown classification algorithms are proposed \cite{RE-OWOD, OCPL, UC-OWOD}. Unlike classification heads, metric learning-based methods try to cluster the instance from the same class or expel the instance from different classes in the latent space. However, such algorithms do not result in better classification accuracy compared with unknown classification heads.

\subsubsection{Catastrophic forgetting}

Similar to other incremental learning tasks, OWOD also needs to face the catastrophic forgetting problem as labeled training data are presented from task to task. Using only new data to fine-tune the model will cause catastrophic forgetting of old classes. Most OWOD methods utilize the exemplar replay strategy to alleviate this. 

Although the implementation of the exemplar replay strategy keeps the accuracy of previously known classes high to some extent, the results still degrade as new labeled data are introduced.

\subsubsection{Unified benchmark and evaluation protocols}

Although many methods follow the standard protocol of the first OWOD method ORE \cite{ORE}, some methods \cite{OW-DETR, RE-OWOD} provide other evaluation metrics and data splits which lead to multiple implementation details. A potential concern with multiple implementation details is determining whether the success of state-of-the-art methods is due to innovative concepts or simply better hyperparameter tuning and improved architecture. Thus, a unified benchmark and standard evaluation protocols need to be used for later methods.

\subsubsection{Differences in Dataset Splits}
Following the first OWOD algorithm ORE \cite{ORE}, most methods report the results under OWOD split, composed of the PASCAL VOC dataset and MS-COCO dataset. However, the potential data leakage from the same super-categories makes the OWOD split not suitable for evaluating the open-world ability of the model. Thus, OW-DETR \cite{OW-DETR} proposes MS-COCO split which separate super-categories in different tasks. Although some later models report the results on both OWOD split and MS-COCO split, other previous methods just report OWOD results only. 

Due to the different training strategies and hardware setups, it is not possible to retrain all the OWOD methods to implement the MS-COCO data split. Thus, later methods are recommended to submit the results on both MS-COCO and OWOD split. Some previous methods can update the MS-COCO results as an appendix.

\subsection{Future Trends}

This subsection briefly discusses some potential future research directions related to Open World Object Detection.

\subsubsection{OWOD with other CV tasks}

Apart from open world object detection, the natural feature of open world settings makes it easy to combine with different research areas, such as semantic segmentation, self-supervised learning, multi-view learning, multi-modality learning, image classification, and so on. The open world scenario is useful for large-scale models as it will automatically detect unknown instances and there is no need to retrain the model using all the datasets, which can save a lot of computational resources. Such combinations need to be investigated.

\subsubsection{Generalization}

In order to generalize open world object detection to multiple application scenarios, the balance between intra-class and inter-class variances is crucial. A high intra-class variance will make the model robust to different instances within a class, meanwhile, it can be detrimental to the unknown detection as the unknown instance are prone to be detected as one of a known-category object. Similarly, the inter-class variance will decide the boundary of different categories which may affect the within-class results, as well as the unknown class detection. Such direction also includes the construction of task-specific dataset.

\subsubsection{Real Application}

Before the real-world application, open world object detection still has many challenges. Firstly, the robustness of such methods needs to be verified, including the robustness towards novel categories, within a class, etc. Second, the inference speed or model efficiency is another key issue, as the computational resource is not always powerful in real applications. Finally, the balance between unknown and known results can be fine-tuned in different application scenarios. \cite{OWODdrive} provided a novel approach for the practical application of OWOD in the field of autonomous driving. 

\subsubsection{Open Vocabulary Object Detection}

Similar to OWOD, the Open Vocabulary Object Detection (OVOD) problem first proposed by Zareian \textit{et al.} \cite{zareian2021open} also tries to go beyond the confined set of base classes labeled in the training stage. Its objective is to identify new classes characterized by an expansive (open) vocabulary during inference. \ym{However, OVOD does not have the ability to learn the object incrementally. Such methods \cite{gu2021open, zang2022open} focus more on unknown detection with the help of large text-vision models. There are also other research works, called OvarNet \cite{chen2023ovarnet}, that try to analyze the gains brought by other information, such as attributes or texts, to recognition or detection.} Thus, OVOD methods are not in the scope of this review paper. As large models offer more avenues for exploration, such directions can be combined together for a real-world object detection solution. 

\subsubsection{Class-agnostic Methods}

According to Table \ref{tab:OWOD}, class-agnostic-based methods achieve a high unknown recall among all the OWOD methods. Such class-agnostic mechanisms need to be investigated further for the specific reasons of high accuracy. In addition, the usage of input-driven pseudo-labeling such as selective search combined with model-driven is a new direction. Selective search as one of the non-data-driven object proposal methods will not be affected by known categories, which meet the need of OWOD. Such methods that do not acquire training data can be further investigated and combined with model-driven region proposal methods.
\section{Conclusion}\label{sec6}

In summary, the open world object detection research area is promising and emerging. Although several related algorithms and methods have been proposed in the past few years, there is still no standard protocol or benchmark to make a fair comparison. The OWOD research still faces many challenges as described above. This paper as the first systematic review of OWOD research provides a comprehensive summary of most OWOD algorithms and methods, evaluation metrics, and commonly used datasets. Moreover, the current challenges and future research directions of OWOD are also analyzed in this paper.

\ifCLASSOPTIONcaptionsoff
  \newpage
\fi



\bibliographystyle{IEEEtran}
\bibliography{./Reference.bib}

\begin{IEEEbiography}[{\includegraphics[width=1in,height=1.25in,clip,keepaspectratio]
{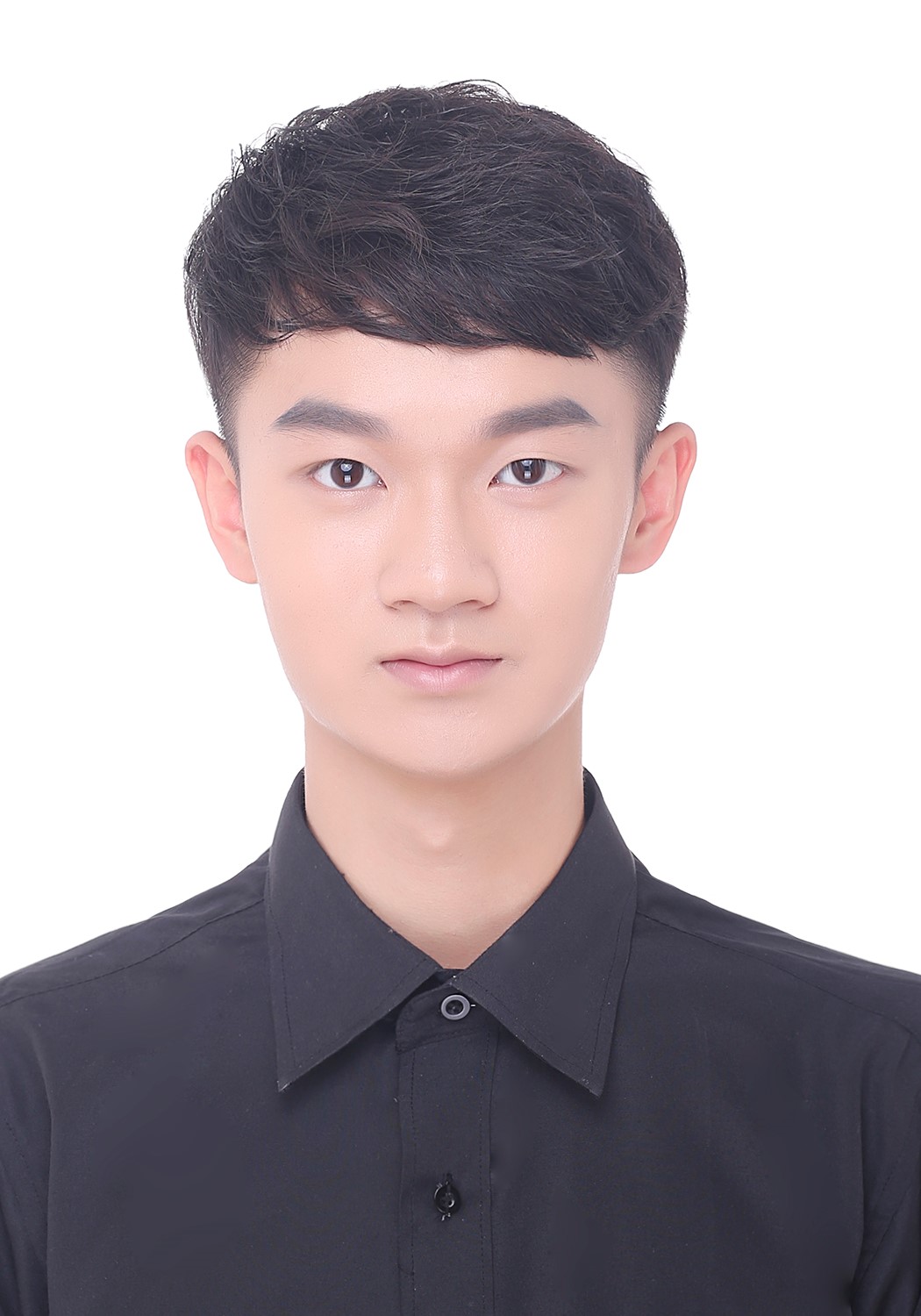}}]{Yiming Li} received his B.Eng. degree from Shanghai University, China, and his M.Sc. degree from Nanyang Technological University. He is currently a Ph.D. student at the School of Electrical and Electronic Engineering, Nanyang Technological University. His research interests include computer vision, object detection, action recognition, and open vocabulary.
\end{IEEEbiography}

\begin{IEEEbiography}[{\includegraphics[width=1in,height=1.25in,clip,keepaspectratio]{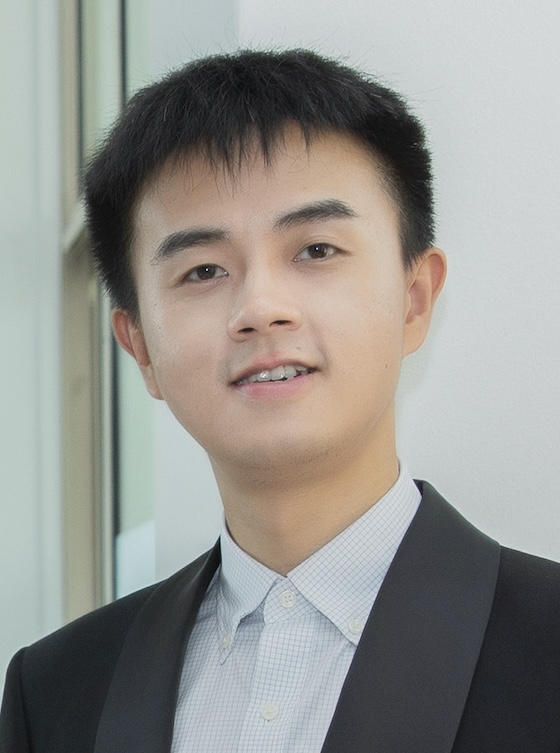}}]{Yi Wang} (Member, IEEE) received the B.Eng. degree in electronic information engineering and the M.Eng. degree in information and signal processing from the School of Electronics and Information, Northwestern Polytechnical University, Xi.an, China,
in 2013 and 2016, respectively, and the Ph.D. degree from the School of Electrical and Electronic Engineering, Nanyang Technological University, Singapore, in 2021. He is now a research assistant professor at the 
Department of Electrical and Electronic Engineering, The Hong Kong Polytechnic University, Hong Kong. His research interests include image restoration, image recognition,
object detection and tracking, and crowd analysis. 

\end{IEEEbiography}

\begin{IEEEbiography}[{\includegraphics[width=1in,height=1.25in,clip,keepaspectratio]{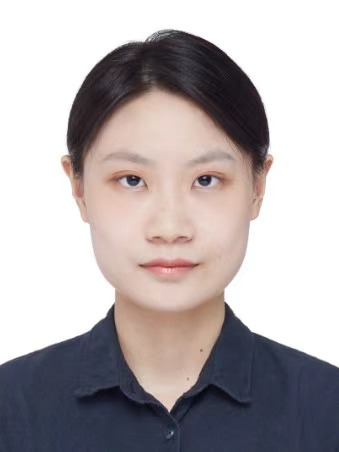}}]{Wenqian Wang}
received her Ph.D. and B.S. degrees from Shandong University, Shandong, China. She is currently a research fellow in School of Electrical \& Electronic Engineering, Nanyang Technological University, Singapore. Her research interests include computer vision, deep learning, anomaly detection and action recognition.
\end{IEEEbiography}

\begin{IEEEbiography}[{\includegraphics[width=1in,height=1.25in,clip,keepaspectratio]{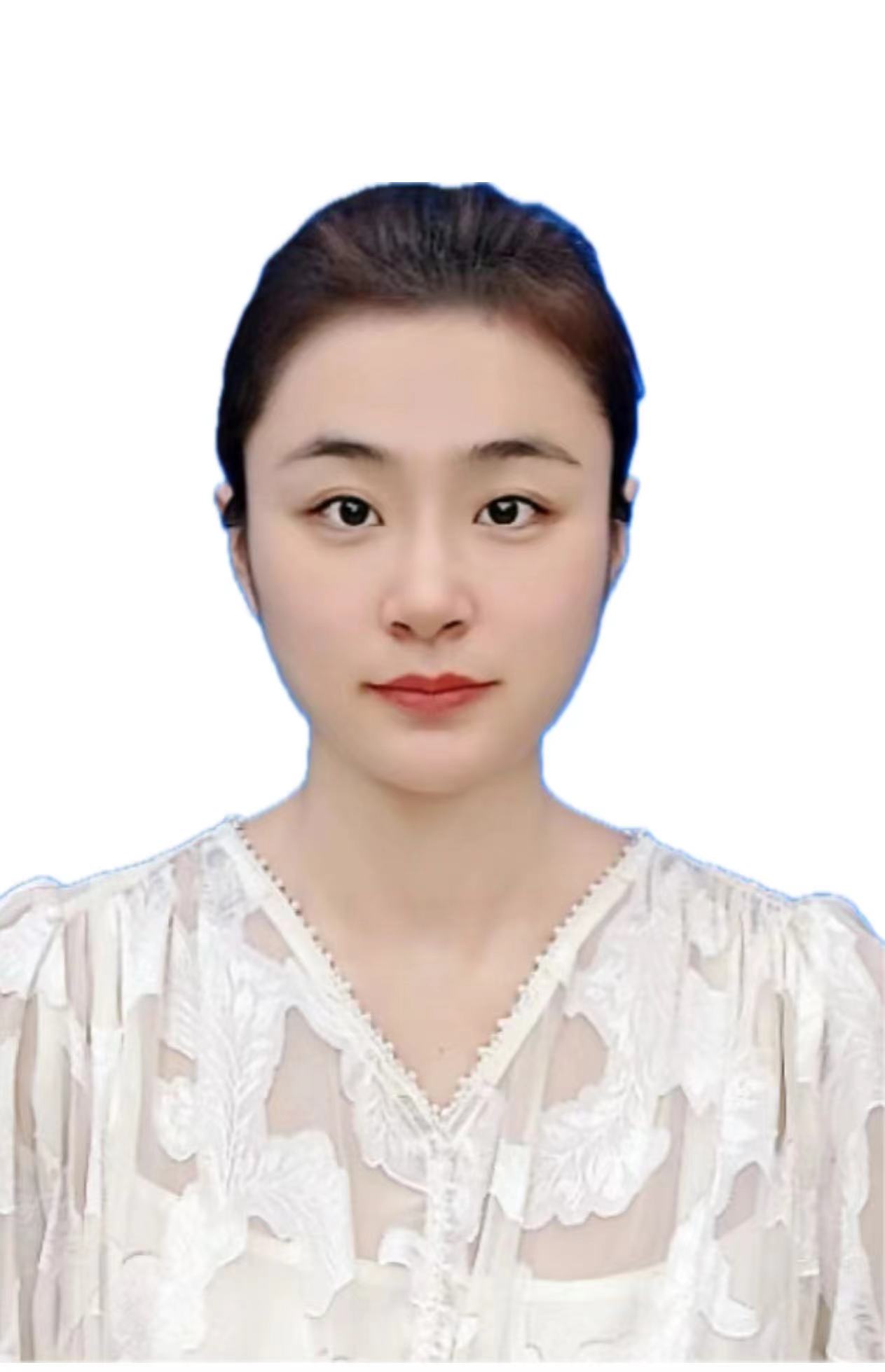}}]{Dan Lin}
received her M.S. and Ph.D. degrees in Software Engineering from Dalian University of Technology, China, in 2017 and 2022, respectively. She is an associate professor in the College of Computer Science and Technology, Harbin Engineering University, Harbin, China. Her research interests include computer vision, and machine learning.
\end{IEEEbiography}
\begin{IEEEbiography}[{\includegraphics[width=1in,height=1.25in,clip,keepaspectratio]{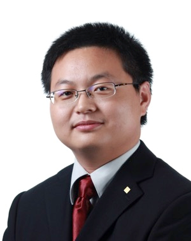}}]{Bingbing Li}
received his B.Eng degrees in Mechanical Engineering from Nanyang Technological University, Singapore in 2011. He is currently a Senior Research Engineer in Continental Automotive Singapore Pte Ltd, and a Ph.D. student at the School of Mechanical and Aerospace Engineering, Nanyang Technological University. His research interests include image recognition, action recognition and their applications in industry.
\end{IEEEbiography}
\begin{IEEEbiography}[{\includegraphics[width=1in,height=1.5in,clip,keepaspectratio]{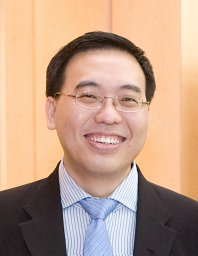}}]{Kim-Hui Yap} (Senior Member, IEEE) received the BEng and Ph.D. degrees in electrical engineering from the University of Sydney, Australia. He is currently an Associate Professor with the School of Electrical and Electronic Engineering, Nanyang Technological University, Singapore. He has authored more than 100 technical publications in various international peer-reviewed journals, conference proceedings, and book chapters. He has also authored a book titled \textit{Adaptive Image Processing: A Computational Intelligence Perspective} (Second Edition, CRCPress). His current research interests include artificial intelligence, data analytics, image/video processing, and computer vision. He has participated in the organization of various international conferences, serving in different capacities, including the technical program co-chair, the finance chair, and the publication chair in these conferences. He was an associate editor and an editorial board member of several international journals.
\end{IEEEbiography}
\end{document}